%% file: main.tex
\documentclass[twoside,11pt]{article}
\usepackage{jair, theapa, rawfonts}

\usepackage{times}
\usepackage{soul}
\usepackage{url}
\usepackage{xcolor}
\usepackage{amsfonts}
\usepackage{mathrsfs}
\usepackage{tikz}
\usepackage{multirow}
\usepackage{longtable}
\usepackage{booktabs}
\usepackage{enumitem}

\usepackage{mathtools}

\usetikzlibrary{positioning,fit,arrows.meta,backgrounds}
\usetikzlibrary{shapes,decorations,arrows}
\usetikzlibrary{calc}
\usetikzlibrary{arrows, automata}

\tikzset{
    module/.style={%
        draw, rounded corners,
        align=center,
        minimum width=#1,
        minimum height=7mm,
        font={\footnotesize}
        },
    module/.default=2cm,
   	beats/.style={thick,->,>=latex,draw}
}

\usepackage{xspace}

\newtheorem{example}{Example}
\newtheorem{definition}{Definition}

\usepackage{color}
\usepackage[textsize=scriptsize]{todonotes}
\definecolor{blue}{RGB}{0, 93, 170}			%Go Big Blue!
\definecolor{darkgreen}{RGB}{0, 102, 0}
\definecolor{stasgreen}{RGB}{0, 152, 0}

\begin{document}

%%
%% The "title" command has an optional parameter,
%% allowing the author to define a "short title" to be used in page headers.

\title{When Is It Acceptable to Break the Rules? Knowledge Representation of Moral Judgement Based on Empirical Data}
%%% Old titles
%When Is It Acceptable to Break the Rules? \\
%Thinking Fast and Slow in Multi-agent Decision Scenarios. \\ Representing Complex Knowledge in Complex Scenarios for Preferences and Moral Judgements}
%Introspection and social norms \\ A Preference-Based Approach}

\author{\name Edmond Awad
\email e.awad@exeter.ac.uk \\
\addr University of Exeter, UK\\
 \AND
\name Sydney Levine
\email smlevine@mit.edu \\
\addr Massachusetts Institute of Technology and Harvard University, USA \\
\AND
\name Andrea Loreggia \email andrea.loreggia@gmail.com\\
\addr European University Institute, Italy \\
\AND
\name Nicholas Mattei 
\email nsmattei@tulane.edu\\
\addr Tulane University, USA\\
\AND
\name Iyad Rahwan 
\email sekrahwan@mpib-berlin.mpg.de \\
\addr Center for Humans \& Machines, Max Planck Institute for Human Development, Germany \\
\AND
\name Francesca Rossi
\email francesca.rossi2@ibm.com\\
\addr IBM Research, USA \\
\AND
\name Kartik Talamadupula 
\email krtalamad@us.ibm.com\\
\addr IBM Research, USA \\
\AND
\name Joshua Tenenbaum
\email jbt@mit.edu\\
\addr Massachusetts Institute of Technology, USA \\
\AND
\name Max Kleiman-Weiner
\email maxhkw@gmail.com\\
\addr Massachusetts Institute of Technology, USA}

\maketitle

\begin{abstract}
%When making decisions humans rely on many kinds of information, inference capabilities, and reasoning processes. We often follow pre-defined rules, that may reflect societal norms when the decision that we are making impacts not only ourselves but others as well. Other times we may decide to challenge these norms and to engage in a more careful and tailored introspection process that evaluates the decision in terms of multiple aspects including the consequences and what others may think, and then decides whether to follow or violate the established norm. Instances of this dual mode of behavior are also outlined in the thinking fast and slow cognitive theory of human reasoning, and the deontology/consequentialist/contractualist approaches to moral judgments.
%Humans make moral judgements about their own actions and the actions of others. Sometimes they make these judgements by employing a complex, possibly utilitarian calculus, other times they may follow simple rules, and yet at other times they may find (or simulate) an agreement among the relevant parties. 

%make these %moral 
%judgements and when they choose to follow pre-defined %rules or rather to engage in an introspection and %consequentialist phase before deciding. This includes when to use a specific moral approach and how to appropriately switch among the various approaches when faced with a decision. 

One of the most remarkable things about the human moral mind is its flexibility.  We can make moral judgments about cases we have never seen before.  We can decide that pre-established rules should be broken.  We can invent novel rules on the fly.  Capturing this flexibility is one of the central challenges in developing AI systems that can interpret and produce human-like moral judgment. This paper details the results of a study of real-world decision makers who judge whether it is acceptable to break a well-established norm: ``no cutting in line.'' We gather data on how human participants judge the acceptability of line-cutting in a range of scenarios. Then, in order to effectively embed these reasoning capabilities into a machine, we propose a method for modeling them using a preference-based structure, which captures a novel modification to standard ``dual process'' theories of moral judgment.

%confronted with the choice of several actions in a suite of hypothetical multi-agent scenarios where they might either follow or break a well-established norm.  

%If we want to build artificial agents that can work effectively with humans, we need to understand how humans switch between these modalities of decision making, and how to embed this knowledge into autonomous agents.

% strugeneralization of CP-nets, a common preference formalism in computer science. We describe a novel extension of CP-nets to both model the scenarios and the moral decisions. We then discuss how our study leads to future research directions in the areas of preference reasoning, planning, and value alignment.
\end{abstract}

% \section{Notes}

% Representation is inspired by the data collection.
% -- Iyad title: When Is It Acceptable to Break the Rules? Knowledge Representation of Moral Judgement based on Empirical Data

% We are doing a study with humans because we want to understand how they thing but we need to validatate and test that.

\section{Introduction}

One of the most remarkable things about the human moral mind is its flexibility: we can make moral judgments about cases we have never seen before.  Yet, on the face of things, morality often seems highly rigid.  Legal codes, employee handbooks, and IRB training manuals all seem to suggest that morality is a system of clearly defined rules.  Indeed, the past few decades of research in moral psychology has revealed that human moral judgment is (at least partially) driven by rules \shortcite{cushman2013action,greene2014moral,holyoak2016deontological,nichols2006moral,mikhail2011elements,levine-preschool-trolley}.  But sometimes, people agree that it is morally appropriate to break the rules.  And sometimes, new rules need to be created when none exist to govern the case at hand.  The field of moral psychology is just now beginning to explore and understand this kind of flexibility -- and there are scant attempts to describe this capacity of the moral mind in computational terms (for a notable exception see \shortcite{levine2020logic}).

Meanwhile, the flexibility of the human moral mind poses a challenge for AI engineers.  AI systems that dynamically navigate the human world will sometimes need to predict and produce human-like moral judgments.  Yet current tools for building AI systems fall short of capturing moral flexibility. In order to build artificial agents that can work effectively with humans, we must equip them with knowledge about how humans make moral judgements \cite{RoMa19}. Hence, AI researchers must engage with the study of human decision makers to understand how they think, and provide a potential formal reasoning methodology to validate and test modeling the results \shortcite{booch2021thinking}.

The goal of this paper is two-fold.  First, we aim to computationally characterize the way that humans both follow and break moral rules in everyday settings, adding to the literature on the computational modeling of moral flexibility.  Second, we aim to capture this pattern of judgments through a novel generalization of a common formalism -- CP-nets \shortcite{cpnets} -- suggesting a way forward for the development of morally competent machines. Note that we consider this work to be a contribution to descriptive ethics, i.e., what people think is morally right or wrong, as opposed to normative ethics, i.e., what is \emph{actually} right or wrong.

Our guide in this project will be dual process theories of moral judgment, which propose that human moral judgment can be understood as a combination of \emph{fast}, heuristic-based thinking and \emph{slow}, deliberative thinking \shortcite{kahneman2011thinking,booch2021thinking}. Our proposal is that the combination of these two modes of thinking enables human moral flexibility, specifically the ability to figure out when pre-established rules should be broken. 

%Often when we make a decision in our daily lives we consider the context, and then we adopt one of two probable approaches: either we follow typical rules or norms that established implicitly or explicitly by the society we live in and we are familiar with (a fast approach), or we may expend a lot of cognitive effort and deliberate about the decision, and then we decide whether to follow the rules or define and adopt new rules for the specific decision scenario. 
%This dual model approach is aligned with the thinking fast and slow theory of D. Kahneman, that describes two broad reasoning approaches: (1) the thinking fast, or system 1, which relies on pre-defined methods and happens almost unconsciously, and (2) the thinking slow, or system 2, that is more deliberate and conscious, and requires all our attention to identify the best course of action \shortcite{kahneman2011thinking,booch2021thinking}.
%
When we follow pre-established moral rules (i.e.: heuristics), we are often using fast thinking.  In contrast, when we are careful and deliberate in our moral decision-making, we are using slow thinking \shortcite{greene2014moral,cushman2013action}.  Up until now, slow thinking has been invoked almost exclusively to explain how people incorporate utility-maximization into their moral judgments.  We propose that slow thinking can also enable moral flexibility; \emph{humans use slow thinking to figure out when the rules should be overridden or when new rules need to be created}.  Moreover, we argue that there are two kinds of slow thinking: outcome-based thinking and agreement-based thinking, which we detail below. Capturing this dual process model of moral flexibility in CP-nets is the engineering challenge that this paper attempts to tackle.  

\section{The Importance of Embedding Ethics Principles in AI Systems}

The idea of teaching machines right from wrong has become an important research topic in both AI \shortcite{yu2018building} and related fields \cite{wallach2008moral}.  Our goal is to formalize and understand how to build AI agents that can act in ways that are morally acceptable to humans \shortcite{RoMa19,LoMaRoVe18,LoMaRoVe18a}.  This challenge has been pursued using a range of computer science approaches, including taking sequences of actions in a reactive environment \shortcite{NBMC19+} and teaching agents how to respond in specific environments \cite{alkoby2019teaching}. Many of these projects address what is called the \emph{value-alignment} problem \shortcite{conf/aaai/ArnoldKS17}, that is, the problem of building machines that behave according to values aligned with human ones \shortcite{russell2015research,loreggia2020modeling,LoMaRoVe18,loreggia2020cpmetric,loreggia2018distance}.

Concerns about the ways in which autonomous decision making systems behave when deployed in the real world are growing in society. Stakeholders worry about AI systems achieving goals in ways that are morally unacceptable to the community impacted by the decisions, also called ``specification gaming'' behaviors \cite{RoMa19}. Thus, there is a growing need to understand how to constrain the actions of an AI system by providing boundaries within which the system must operate. To tackle this problem, it is useful to take inspiration from humans, who often constrain the decisions and actions they take according to a number of exogenous priorities, be they moral, ethical, religious, or business values \cite{Sen,LoMaRoVe18,LoMaRoVe18a}, and we may want the systems we build to be restricted in their actions by similar principles \cite{conf/aaai/ArnoldKS17}. The overriding concern is that the agents we construct may not obey these values while maximizing some objective function\footnote{https://www.wired.com/story/when-bots-teach-themselves-to-cheat/ - May, 31st 2021} \cite{RoMa19}.
But humans are also experts at figuring out when the constraints can and should be broken.  This is a hallmark of human moral flexibility -- humans can figure out when simple rules should be overridden.  Can we enable AI to do the same?

%While giving a machine a code of morals or ethics is important, there is still the question of \emph{how to provide the behavioral constraints to the agent}.  
There are two popular techniques to encoding constraints into AI systems: bottom-up and top-down \cite{wallach2008moral}.  A bottom-up approach involves teaching a machine what is right and wrong by providing examples of ``correct'' decisions, e.g., providing a reinforcement learning system with demonstrations from humans in a setting such as self driving cars or exploring mazes \shortcite{allen2005artificial,balakrishnan2018incorporating,glazier2021making}. In top-down approaches, behavioral guidelines are specified by explicit rules or constraints on the decisions space, e.g., by making a rule that a self-driving car must never strike a human \cite{svegliato2021ethically}.  However, both of these kinds of models will struggle to determine when constraints should be overridden. Top-down approaches will err when the rules or constraints on the system are too general to deal with a particular edge case or unusual circumstance. Bottom-up approaches will err when a case that should be an exception to a rule differs dramatically from anything in the training set.

In this paper, we describe cases where humans think it is acceptable to violate a rule.  We then propose a formalism that will help the AI community move forward in encoding this kind of moral flexibility into AI systems.

\section{Fast and Slow Thinking Guide Moral Judgment}

\subsection{Brief Overview of Previous Work}

Many human decision-making processes can be characterized by the System 1/System 2 model of cognition, often known as the ``thinking fast and slow'' approach, \cite{kahneman2011thinking}. This model describes two broad reasoning approaches: (1) thinking fast, or System 1, which relies on predefined heuristics or rules, and (2) thinking slow, or System 2, that is more deliberate.  Each of these systems has their merits.  System 1 thinking is efficient, requiring only limited computational resources and contextual information.  When System 1 heuristics are deployed in the environments they were intended for, they often produce good-enough decisions; though they sometimes can lead to sub-optimal choices when deployed in edge cases.  System 2 decision-processes, on the other hand, are cognitively intensive and allow the decision-maker to flexibly integrate information from disparate sources, leading to decisions that can be carefully tailored to the case at hand.  These decision-processes can be used in combination in a way that optimizes payoffs given the limitations of the computational resources available \cite{lieder2020resource,simon1955behavioral,simon1956rational}. 

%Often when we make a decision in our daily lives we consider the context, and then we adopt one of two probable approaches: either we follow typical rules or norms that established implicitly or explicitly by the society we live in, often without questioning them, or we evaluate the consequences of the decision and then we decide whether to follow the rules or define and adopt new rules for the specific decision scenario.

The System 1/System 2 model of decision-making has a well-known analogue in the moral psychology literature, known the ``dual process'' model of moral judgment \cite{cushman2013action,greene2014moral,kleiman2015inference,holyoak2016deontological}.  Dual process models argue that we make moral judgments using a combination of heuristic-like rules (System 1-style reasoning) and more deliberative System 2 methods.  The dual process models draw their inspiration from two of the major branches of moral philosophy.  The rule-based System 1 is inspired by \emph{deontological} theories of moral philosophy, which focus on constraints on actions.  The deliberative System 2 is typically associated with \emph{consequentialist} theories, which focus on maximizing the utility of outcomes. System 1 applies a hard and fast rule and does not consider the subtleties or complexities of the current scenario.  In contrast, System 2 acts more like a utilitarian reasoner, it attempts to quantify utility losses and gains, which enable it to make a judgment.  For example, when Bob refrains from taking something he cannot afford from a store even though he really wants it, he may be following an intuitive moral rule: no stealing.  When Susan decides to donate \$5 to buy mosquito nets rather than chocolate bars for children in crisis, she may be using System 2 reasoning and comparing the values of the outcomes of the candidate actions. 

\subsection{A New Kind of Slow Thinking: Contractualism}
We mentioned above that the central ideas of two of the main branches of moral philosophy (rules and outcomes) are represented in theories of moral psychology.  Curiously, the central idea of the third major family of philosophical views -- \emph{contractualism} -- has been virtually absent from theories of moral psychology.  Contractualist views ask us to consider what agreements could be adopted that would lead to mutual benefit \cite{gauthier1986morals,rawls1971theory,scanlon1998we,habermas1990moral}. Recent work has begun to show that, despite being neglected for so long, contractualist mechanisms play a critical role in the moral mind \shortcite{levine2018contractualism,levine2020logic,baumard2013mutualistic}.  

The specific agreement-based process we focus on in this paper is \emph{universalization}, which is a way of making a moral decision by asking ``what if everyone felt at liberty to do that?'' \cite{levine2020logic}.  When people universalize, they imagine a hypothetical world where everyone is allowed to act in a certain way.  If things go well in that hypothetical world, the action in question should be allowed.  If things go badly, the action should be prohibited.  This counts as agreement-based because it is a way of asking if any person is taking a special privilege for themselves that they couldn't grant to everyone.  If everyone could feel free to do the action, then presumably mutual agreement could be reached that the action is permitted.
Universalization is likely to be a System 2 process because running the universalization computation is resource-intensive and requires a lot of information about the particular decision-making context \cite{levine2020logic}.  However, more empirical work needs to be done to be certain that this is the case.

\subsection{Fast and Slow Thinking Explain Moral Flexibility}

We propose that System 1 processes enables rule-following behavior, as is widely agreed upon in the literature. In contrast, \textit{System 2 processes can enable moral flexibility}.  Specifically, humans use System 2-thinking to figure out when the rules should be overridden.  Moreover, we argue that there are two kinds of System 2 thinking: outcome-based thinking and agreement-based thinking.

Sometimes people decide that it is morally acceptable to break previously established rules because doing so would bring about a better outcome than following the rule \cite{greene2014moral,hare1981moral}.  In general, there is a rule against non-consensual harmful contact, i.e., no hitting. But if pushing someone out of the way of a speeding train could save their life, then it is morally permissible to do so.
Yet in other cases, people may decide that it is morally acceptable to break a rule because the person whom the rule protects would agree to have the rule violated \cite{levine2018contractualism}.  There is rule against taking things that don't belong to you (or: no stealing), but it may be OK to take coffee grounds from your co-worker's desk if you know they would consent to it, given that they are likely to be caught in a similar situation in the future and want to take some coffee from you.

In sum, we propose that both contractualist (agreement-based) and consequentialist (outcome-based) System 2 reasoning help humans figure out when it is acceptable to override the rules.  We aim to be able to capture this computationally, as described in Section \ref{sec:comp_approach} below.

%We incorporate contractualist reasoning into our model of System 2 moral judgment alongside consequentialist reasoning.  %Combined with System 1 rule-following, these three systems constitute a Psychological ``Triple Theory''.

\subsection{Towards a Psychological Triple Theory}
There have been various attempts by philosophers to unite the three major threads of moral philosophy (outcomes, rules, and agreement) into a single unified view \cite{parfit2011matters,hare1981moral}. Parfit called his unified view a ``Triple Theory''.  To date, no psychological theory has attempted to explain how rules, outcomes, and agreement are all integrated in the moral mind.  Our work is one step on the way to creating a Psychological Triple Theory.

\section{Our Paradigm: Waiting in Line}
Our paradigm involves asking our subjects when it is morally acceptable to break simple and generally agreed upon rules. We depart from the typical work in this area which has often focused on high-stakes judgments in somewhat uncommon scenarios, e.g., a runaway trolley headed towards innocents \shortcite{awad2018moral}, and we rather probe people's moral intuitions in a very common everyday scenarios: standing in line to receive a service. Intuitively, it seems like an easy feat to figure out how to wait in line. In fact, one simple widely agreed rule governs the process of waiting in line: each person in line is helped in the order that they arrive (first-in-first-out). If true, navigating a situation that requires waiting in line would simply involve getting in the back of the line and waiting your turn.  

However, a few moments' reflection reveals that we can intuitively evaluate all kinds of exceptions to this seemingly simple rule about waiting in line. For example, say that you are in a deli and have waited in line and ordered a bowl of soup, but just as you are about to begin eating the soup, your spoon falls on the floor.  It is probably acceptable for you to get a new spoon without waiting in line.  Or say that you just want a glass of tap-water.  Here, too, usually you are allowed to cut to the front of the line without waiting.  Or if you are assisting someone who has just fallen off a bike outside, you can probably purchase a bottle of water without waiting in line. On the other hand, it is probably not acceptable to cut to the front of the line to order a soda, even though that might delay the line just as much as the person requesting tap water.  How do humans figure out when it is acceptable to cut in line and when it's not? And what kind of reasoning do people use in making their judgments?  We propose that System 2 reasoning, which typically involves utility calculations and/or agreement-based reasoning, helps subjects figure out when it is OK to break the rules.
%Despite first appearances, figuring out how to wait in line cannot be governed by simple rules.  %In this paper, we show that subjects make decisions about when it is acceptable to cut in line by considering what everyone in the situation \textit{would agree to}.  

%\loreg{Waiting in line scenario provides an interesting test-case to study how people make normative judgments.  As highlighted before, this scenario sits at the interesting intersection between moral and conventional rules. Queuing after all is entirely learned and rules about it are dependent on the social environment. However, those rules have a moral goal, namely, to distribute resources efficiently and fairly, so violating the convention of the line is morally problematic because of the implications for fairness.}

We use three different contexts for waiting in line: a line at a deli, at an airport, and at a bathroom.  In each context, we present participants with a series of reasons why someone may want to cut the line. We ask subjects to judge whether it is acceptable to cut in line in the cases described.  We also ask subjects to evaluate each scenario on a series of other measures, which tap into the underlying cognitive mechanisms that are driving their moral judgments. For instance, we ask subjects to estimate how long the cutter would delay the line, the benefit to the cutter, the detriment to the line, and so forth. Using these metrics, we can generate an expected utility calculation for the action. We also ask subjects what would happen if this type of line-cutting were always allowed, a proxy for whether everyone would agree to allow this person to cut \cite{levine2020logic}.  We also code each of these cases for obvious instances of a rule violation, where the rule about cutting in line is construed in simple terms, e.g., you must wait in line at a deli if you intend to buy something. Full experimental details are provided in Section \ref{sec:exp_details}.

Each scenario is modelled in our computational framework through a set of \emph{scenario variables}, such as the location or the reason to ask to cut the line, and gives rise to a number of descriptive \emph{evaluation variables} dependent on the scenario variables. We then model this data in a preference-based framework that links scenarios, evaluation variables, and moral judgement, claiming it can serve as a central modeling and reasoning framework for human moral reasoning and decision making.

\section{Our Computational Approach}\label{sec:comp_approach}

\subsection{A Computational Approach to Moral Judgment is Critical for AI Development} The vast majority of work characterizing human moral decision-making, both in psychology \cite{doris2010moral,haidt2007new} and experimental philosophy \cite{knobe2007experimental,alexander2012experimental} has focused on identifying factors that are relevant to moral judgment (e.g., affect, rules, utility calculations). Our proposal contributes to an emerging body of work that goes beyond simply identifying these factors, but also seeks to characterize \emph{in computational terms} the mechanisms underlying moral judgments \shortcite{levine2020logic,kleiman2017learning,kim2018computational,van2019computational,engelmann2022weigh,jiang2021delphi}.  Describing the mind in computational terms is a critical step in building AI that can produce and interpret human moral judgment \shortcite{kleiman2017learning,bonnefon2020moral,russell2019human}. Moreover, our work goes even further and describes the way that current methods of implementing morality in AI systems would need to be modified to capture the computational mechanisms we describe.

\subsection{CP-Nets: A Preference-Based Approach}

%We employ two computational modeling tools: a preference-based framework %(generalized) CP-nets and probabilistic planning.

Decisions and actions are linked to the concept of preferences:
we choose one among all possible decisions because we prefer the resulting state-of-affairs over the ones generated by the other decisions. The issue of modelling and reasoning with preferences in an artificial agent has been the subject of a very active research area for many years. Several frameworks have been defined, and their properties studied, for example related to expressivity, computational complexity, and easiness of preference elicitation \cite{rossi2019preferences}. The centrality of preferences is true also in the case of moral judgement: we consider a decision more morally acceptable than another one if its impact on others is preferred according to our moral values \cite{Sen,harsanyi1977morality}. Finding a  way to model and reason with such values, and corresponding preferences, is central to build artificial agents that behave in a way that is aligned to humans values \cite{LoMaRoVe18}. 

%\loreg{We can think of the decision-making process as an evaluation of the resulting state-of-affairs. 
%We choose one among all the available options because we prefer the resulting state-of-affair over the others. Thus,  we can think of
% According to Sen \cite{Sen}, moral judgements as a form of preferences, driven by moral reasoning \cite{Sen,harsanyi1977morality}. Finding a feasible way to model and reason with these preferences are central in the field of Artificial Intelligence in order to ensure a convergence of AI agents behavior with humans values. Indeed, the issue of modelling and reasoning with preferences in an AI system has been the subject of a very active research area for many years, and has produced many frameworks to deal with preferences and embed them into an AI decision making system. Different frameworks differ on properties related, for example, to expressivity, computational complexity, and easiness of preference elicitation \cite{rossi2019preferences}.

In this work, we specifically investigate the conditions under which humans decide to break previously established rules in multi-agent scenarios. We then use a preference-based framework to 
%or moral decisions and judgments using studies of real-world decision makers. \loreg{We also discuss one potential way to leverage our results 
to model and embed the underlying processes they use into a machine. 
%In this regard, we propose a new generalization (that we  call SEP CP-net) of the well known preference framework  CP-net.}
%
%
We exploit an existing preference modelling and reasoning framework and we generalize it to include also scenario and evaluation variables. Conditional Preference networks (CP-nets) are a graphical model for compactly representing conditional and qualitative preferences \cite{cpnets}. CP-nets are comprised of sets of {\em ceteris paribus} preference statements (cp-statements). For instance, the cp-statement, {\em ``I prefer red wine to white wine if meat is served,"} asserts that, given two meals that differ {\em only} in the kind of wine served {\em and} both containing meat, the meal with red wine is preferable to the meal with white wine.  CP-nets have been extensively used in the preference reasoning, preference learning, and social choice literature as a formalism for working with qualitative preferences \shortcite{DHKP11a,RVW11a,BCELP16a}. CP-nets have even been used to compose web services \shortcite{wang2009web} and other decision aid systems \shortcite{pu2011usability}. While there are many formalisms to choose from when modeling preferences, we focus on CP-nets as they are graphical and intuitive.  Transforming the results of our experiments into other formalisms is an interesting and important direction for future work.

\begin{table}\label{tab:reason}
\centering
{\small
%\resizebox{0.97\linewidth}{!}{
\begin{tabular}{ lp{0.65\textwidth}p{1cm}p{1cm}}
 Label & Scenario Description & Main Func. & Already Waited\\
 \hline
 \textbf{Deli Scenarios}\\
 \hline
 Spoon & A customer who is eating soup at the deli dropped his spoon on the floor and needs another one. & False & True \\
 Water & A customer who is eating lunch at the deli wants more a refill on tap water. & False & True \\
 Soda & A customer who is eating lunch at the deli wants to buy another soda. & True & True \\
 Catering Order & A customer wants to ask a series of questions about a catering order that he will pick up next week. & False & False\\
 Fasted & A customer walks in who has just finished fasting for 24 hours for a colonoscopy and is extremely hungry. & True & False\\
 Diabetic & A customer walks in who is diabetic and urgently needs sugar. & True & False\\
 Oven Repair & %The oven in the deli broke this morning and the deli therefore removed half the items from the menu. 
 The oven-repair technician shows up and needs to ask the cashier a series of questions about the oven so he can fix it. & False & False\\
 Soap & A customer uses the last of the hand soap in the bathroom. & False & False\\
 Toilet Paper & A customer notices that the bathroom is out of toilet paper. & False & False\\
 Spouse & A customer walks in who is married to a customer who is currently placing an order with the cashier. & True & False\\
 Father & The father of the family is currently placing an order with the cashier. & True & False\\
 Sandwich & A customer walks in who wants to order a sandwich. & True & False  \\
 \hline
 \textbf{Bathroom Scenarios}\\
 \hline
 Wash Hands & Someone at the back of the line just needs to wash their hands. & False & False\\
 Cleaner & Someone arrives who needs to clean the bathroom. & False & False \\
 Vomit & Someone at the back of the line needs to throw up immediately. & True & False \\
 Get Jacket & Someone at the back of the line thinks they forgot their jacket in the bathroom. & False & True\\
 Friend & Someone at the back of the line is a friend of someone at the front of the line. & True & False\\
 Aid & Someone at the back of the line is an aid to an elderly person at the front of the line. & True & False\\
 Use Bathroom & Someone at the back of the line needs to use the bathroom. & True & False\\
 \hline
 \textbf{Airport Scenarios}\\
 \hline
 Departure in 20min & The flight leaves in 20 minutes. & True & False \\
 Crying Baby & Standing with a baby who is crying very loudly. & True & False\\
 Forgot Jacket & Forgot their jacket at the check-in counter. & False & True\\
 Cafe Worker & Works at a cafe inside the airport. & False & False\\
 Go to Bathroom & Has to leave the line to go to the bathroom. & True & True \\
 Departure in 3h & The flight leaves in 3 hours. & True & False \\
\hline
\end{tabular}
%}
}
\caption{Description of all 25 scenarios used in our experiments.}
\label{tab:scenarios}
\end{table}

\begin{table}[]\label{tab:questions}
{\small
\centering
\begin{tabular}{p{0.25\textwidth} p{0.75\textwidth}}
\toprule
Variable              & Prompt  \\ \midrule
\textbf{Global Welfare} & Think  about  the  well-being  of  all  the  people  in  line  combined.  How  are  they affected by the person cutting in line? \\ \addlinespace[0.5em]
\textbf{First Person Welfare}  & How much worse off/better off is the first person in line?  \\ \addlinespace[0.5em]
\textbf{Middle Person Welfare} & How much worse off/better off is a person standing in the middle of the line?   \\ \addlinespace[0.5em]
\textbf{Last Person Welfare}   & How much worse off/better off is the last person in line?     \\ \addlinespace[0.5em]
\textbf{Line Cutter Welfare}   & How much worse off/better off is the person that cut in line?  \\ \addlinespace[0.5em]
\textbf{Universalization}      & Think about the person who cut in line. How much worse off/better off would it be for people who come to the deli if everyone who was in this situation cut in line? \\ 
\textbf{Likelihood} & On any given day, how likely is it that this scenario going to happen? \\ \addlinespace[0.5em]
\textbf{Judgement} & Is it acceptable to cut the line? (yes or no). \\
\bottomrule
\end{tabular}
}
\caption{Questions that each respondent was asked in order to evaluate each possible scenario.}
\end{table}

\section{Experimental Details}
\label{sec:exp_details}

To gain insight into how humans make moral judgements we ran an experimental study on Amazon MTurk. Informed consent was given by all participants.  This study was approved by the Massachusetts Institute of Technology IRB. 407 subjects participated in the study.  Following attention checks, the data from 301 subjects was retained for analysis. Subjects were randomly assigned to one of three story contexts: 
\begin{description}[itemsep=0em]
\item[Deli.] waiting in line at a deli (12 scenarios); 
\item [Bathroom.] waiting in line for the bathroom (7 scenarios); 
\item[Airport.] waiting in line at the airport security screening (6 scenarios). 
\end{description}
Participants were asked to read a set of short vignettes about people who wanted to cut in line for a range of different reasons. For example: ``Imagine that there are five people who are waiting in line at a deli to order sandwiches for lunch. There is only one person (the cashier) working at the deli. A customer who is eating soup at the deli dropped his spoon on the floor and needs another one.  Is it OK for that person to ask the cashier for a new spoon without waiting in line?'' Subjects were allowed to answer ``yes'' or ``no''.  Complete descriptions of the scenarios are given in Table \ref{tab:scenarios}.

Subjects were also asked to evaluate each scenario on a range of different metrics which might correspond to the cognitive mechanisms underlying their moral judgments.  The full list of evaluation questions is given in Table \ref{tab:questions}.  Subjects responded on a scale of -50 (a lot worse off), 0 (not affected), +50 (a lot better off); likelihood was judged between 0 (not likely) and 100 (very likely).

Half the subjects were shown the evaluation questions for all the scenarios followed by the permissibility questions for all the scenarios; the other half of the subjects received the blocks of questions in the reverse order.  This was designed to test, and eliminate if necessary, the effects of evaluation then judgement versus judgement then evaluation.

\begin{table}[]\label{tab:scenario}
{\small
\centering
\begin{tabular}{p{0.25\textwidth} p{0.75\textwidth}}
\toprule
Variable              & Scenario Description  \\ \midrule
\textbf{Reason} & The particular reason for cutting the line, see Table \ref{tab:reason}. \\ \addlinespace[0.5em]
\textbf{Location}  & The particular location of the scenario: \emph{Deli}, \emph{Airport}, or \emph{}.  \\ \addlinespace[0.5em]
\textbf{Main Service}   & Whether or not the individual cutting the line had the goal of accessing the main service. \\ \addlinespace[0.5em]
\bottomrule
\end{tabular}
}
\caption{Variables that went in to constructing each scenario.}
\end{table}

Finally, we coded each vignette for the presence of a possible confounding feature: whether or not the person attempting to cut had the goal of accessing the main service the line was providing. The main service for the deli line was the sale of an item, for the airport scenario it was security screening, and for the bathroom it was the use of the toilet. Note that defining what counts as the ``main service'' being provided to the line is open to interpretation.  We think that variation in this interpretation likely impacts subjects' judgments about the acceptability of cutting in line.  For instance, one might view the main service of the deli line as receiving anything from the cashier (rather than purchasing something). Our characterizations of the main function of the line are rough approximations meant to describe one view that seems commonly held. These coding can be found in Table \ref{tab:scenarios}.

%two features: (1) whether the person attempting to cut in line had already waited in line and (2) whether the person attempting to cut had the goal of accessing the main service the line was providing.  

At the end of the survey, participants were given an attention check as follows: ``Finally, we are interested in learning some facts about you to see how our survey respondents answer questions differently from each other.  This is an attention check.  If you are reading this, please do not answer this question (do not check any of the boxes).  Instead, in the box below labeled `Other' please write `I am paying attention'.  Thanks very much!''  This was followed with a list of levels of education and a free-response box labeled ``Other.''  Participants who checked any box or failed to write ''I am paying attention'' in the appropriate place were screened out of the study.

\begin{figure}
\centering
\includegraphics[width=0.85\linewidth]{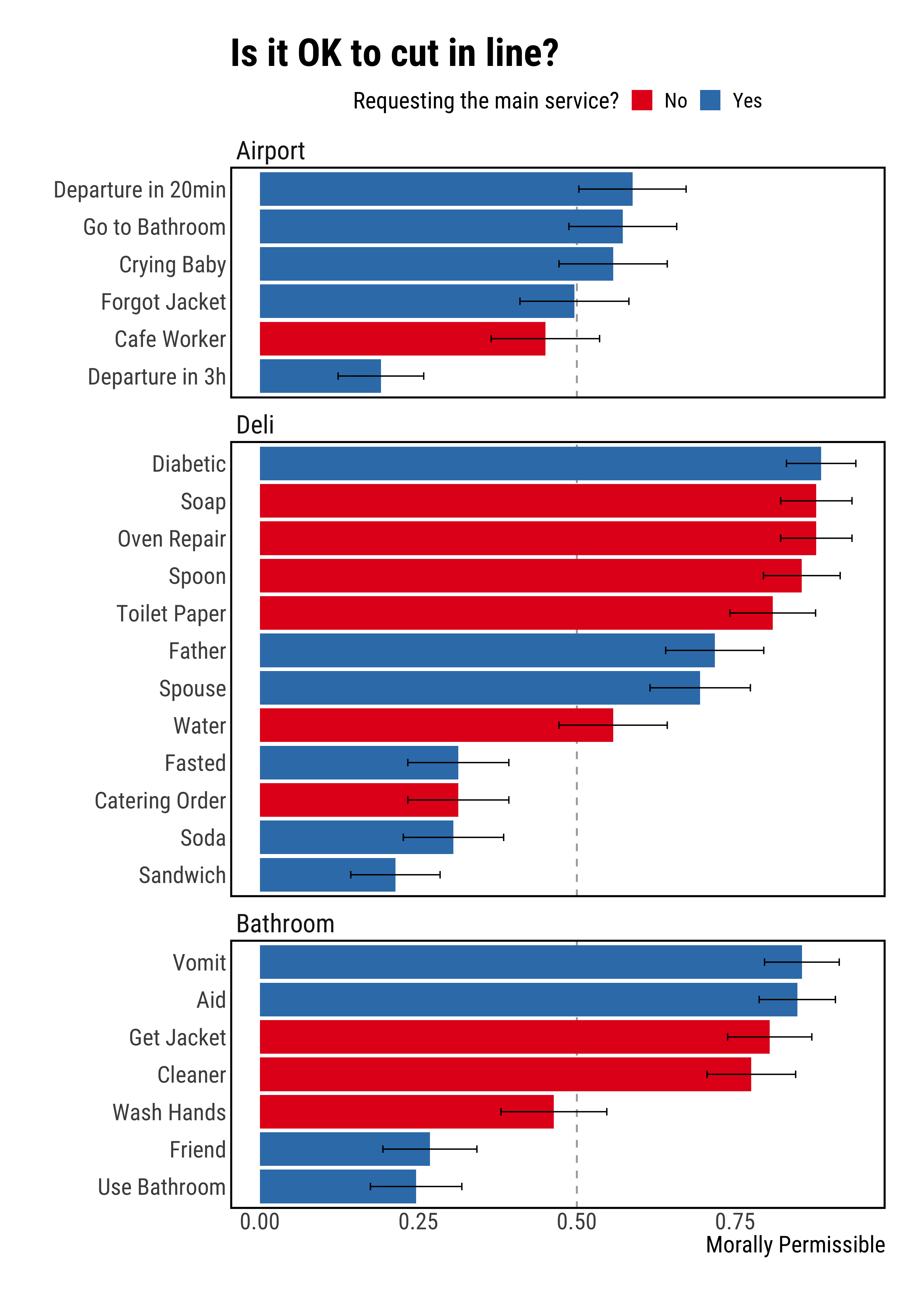}
\caption{
Moral judgements about cutting in line in each of the scenarios. Color indicates if the person cutting in line is requesting the main service or not (blue for Yes, red for No). Error bars are $95\%$ confidence intervals. As we can see, a simple rule, such as ``it is ok to cut if you are not requesting the main service'' is not sufficient to explain variation.  %Figure (a) shows the moral permissibility of cutting in line in each of the scenarios in Study 1, while
}
\label{fig:bars_study2}
\end{figure}

\section{Data Analysis} \label{sec:eda}

\subsection{Moral Judgments}

\begin{figure}
  \centering
\includegraphics[width=0.9\columnwidth]{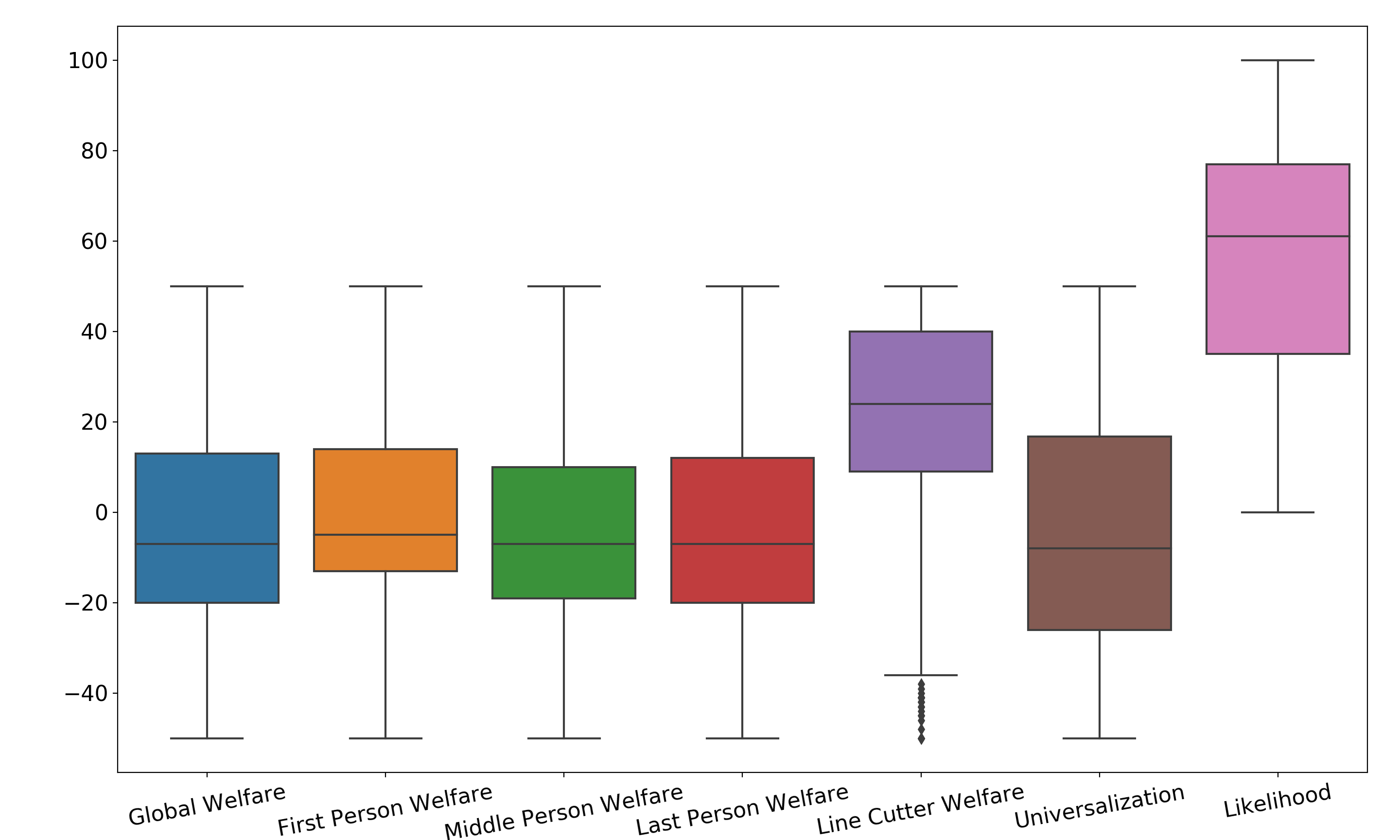}
  \caption{Box-plots depicting the distribution of responses to the evaluation questions, %in Study 1 (top) and Study 2 (bottom), 
  polled across scenarios.  %In general, the cost to all members of the line have similar distributions but that the spread in Study 2 is wider than in Study 1.
  }
  \label{fig:boxplot}
\end{figure}

%We turn first to our subjects in Study 1. 
Figure \ref{fig:bars_study2} shows overall moral permissibility data (i.e., ok or not-ok judgment) for each of the scenarios. The most notable feature of this data is that moral permissibility is graded; cutting the line is endorsed probabilistically by our subjects rather than in an all-or-none fashion.  This is the first hint that our subjects are not using a simple rule to figure out when it is permissible to cut in line.  A rule like ``don't cut'' for instance, would produce unanimously low permissibility for all cases.  

A slightly more sophisticated version of this rule-based approach would be that subjects use the rule ``don't cut'' but realize that the rule is not operative in certain scenarios.  This would yield a slightly more complex rule such as ``cutting is allowed only when you are not requesting the main service.''  (See Figure \ref{fig:bars_study2}.) This would yield a binary all-or-none pattern of results, with some instances of cutting being permissible and some not.  Instead, it seems that a more sophisticated understanding of the computations behind subjects' moral judgments is required. The aim of our model is to be able to predict our participants' graded permissibility judgments.  

%We repeat the same analysis for Study 2 in Figure \ref{fig:bars_study2}.  Comparing these two we can see that the pattern of responses between Study 1 and Study 2 are generally the same, though the participants in Study 2 are generally more morally permissive. %Hence, separating the groups into an evaluation and judgement did not significantly change the results.

%\francesca{\bf FRANCESCA: so what do we say about system 1 vs system 2 here, given that the two studies give similar results?}
%\loreg{We conjecture that people use a two-system fashion to make a decision: in some scenarios, people might make a decision based on few information (e.g., just the location) and in other scenarios a more deep introspection seems to be required in order to evaluate the situation and decide the preferred consequence. %This is also justified by the fact that people reply faster to some scenarios than others.
%}

%\sydney{Feel free to ignore for this submission, but comment for the future: Seems like we should have a graph for Study 2 moral permissibility as well, or at least a comment on whether the scenarios produce broadly the same pattern of data?  It also seems fine to entirely drop Study 1.  Is it doing anything for us that Study 2 doesn't do?}

%Turning to Study 2, 
\subsection{Checking for Order Effects}
First, we checked if order (whether subjects were asked the evaluation questions first and then asked to make a moral judgement, or the opposite) had any effect on moral judgments.  To test this we ran a Wilcoxon signed-rank test against the null hypothesis that \emph{changing the order of evaluation and judgement does not influence the judgement value}.  We ran this for all 25 scenarios described in Table \ref{tab:scenarios}. %The full set of $p$-values is available in the supplemental material. 
The test did not reject the null hypothesis for any of the 25 scenarios. Hence, under the conditions of this study, asking individuals to think closely about the evaluation questions did not cause them to change their judgements. 
%\francesca{\bf So? what are the implications on system 1 vs system 2?}
%This seems to respect the internal deliberation process that individual experience to make a decision. 
We conjecture that this happens because people already went through the internal evaluation process when they made a decision. Thus asking for it before or after does not influence the decision they made.  We therefore pooled subjects in both order conditions for the remaining analyses.
%Throughout the remainder of this work we %combine use results from Study 2.
%\francesca{\bf What does it mean to combine the participants of study 2?}

\begin{figure}
  \centering
    \includegraphics[width=1.0\columnwidth]{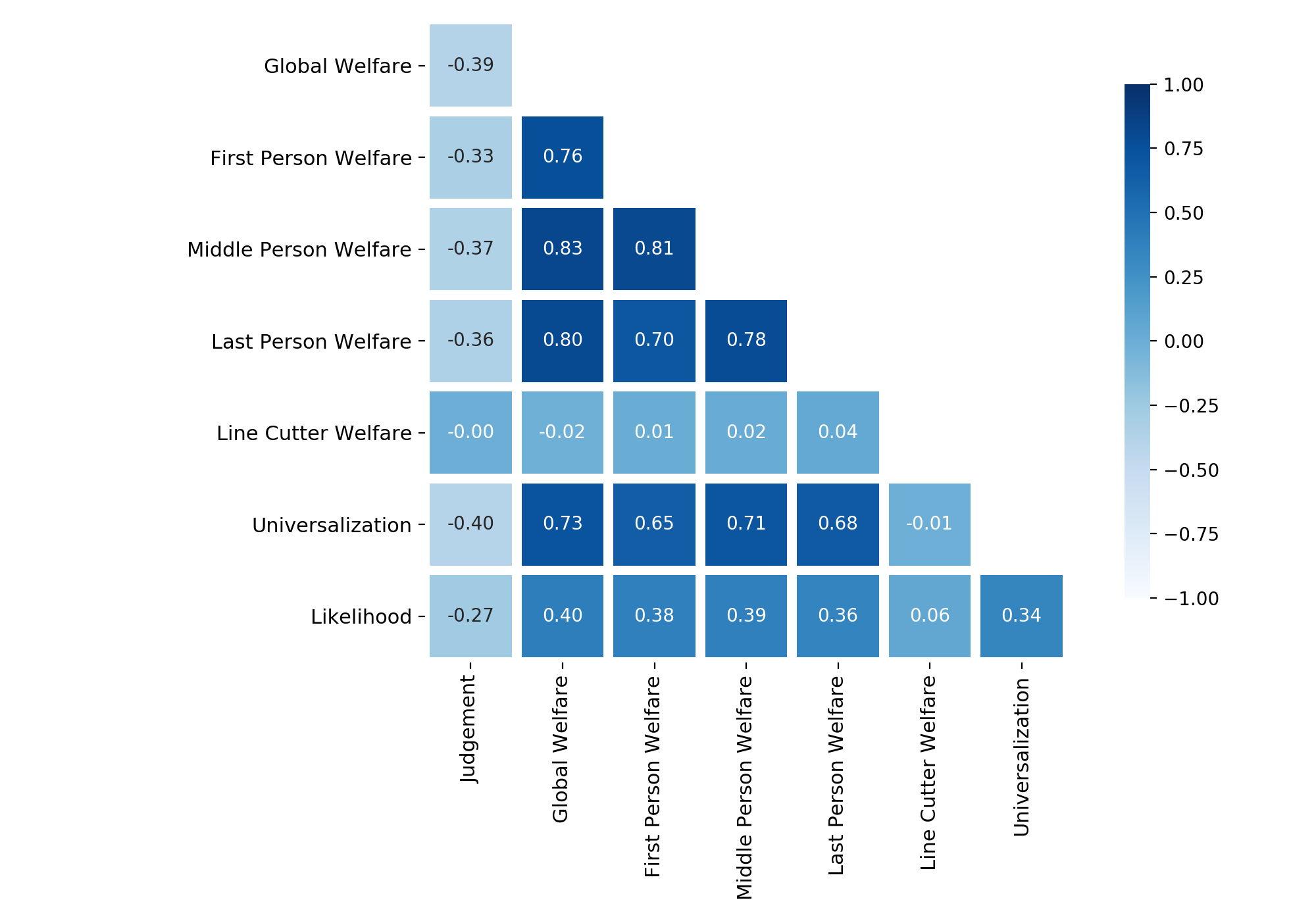}
\caption{Cross-correlation matrix for all scenarios.
The moral judgement is labeled as PREF and is negatively correlated with all the evaluating metrics, indicating that as any measure gets worse, the moral permissibility goes down.}
\label{fig:corr_mat}
\end{figure}

\subsection{Evaluation Questions}
Figure \ref{fig:boxplot} depicts the distribution of responses by all participants for the evaluation questions.% in Study 1 (top) and Study 2 (bottom).  While generally these are very similar between study groups, the inter-quartile range (spread between the 25\% and 75\% response) of the participants in Study 2 is significantly wider. 
We conjectured that the different evaluation variables might draw out importantly different elements of the scenarios, allowing for the emergence of a better predictor by using multiple variables. For instance, the well-being of the first person could matter most because she has the most to lose if someone cuts in line, i.e., her change in wait time is proportionally greatest. On the other hand, it might emerge that the last person is most important because waiting additional time might actually make it no longer worthwhile to wait at all. Furthermore, \emph{Global Welfare} or \emph{Middle Person Welfare} might be the best because they provide an aggregate estimate or average estimate. As it turns out, people do not judge these metrics differently (See Fig.\ref{fig:boxplot}).  Additionally, there is a fair amount of negative skew for the question \emph{Line Cutter Welfare}, indicating that some participants felt that even if cutting the line was allowed, they were not receiving much benefit.

\subsection{Relationship Between Evaluation Questions and Moral Judgment}
Finally, we wanted to see if there were any strong correlations between the various evaluation questions and the moral judgement. Figure \ref{fig:corr_mat} shows the cross-correlation between all responses from the subjects. As one might expect, the questions about the individuals in line are highly correlated, further indicating that most subjects respond to these questions with similar evaluations.  In addition, as predicted, all evaluation variables are negatively correlated with the moral judgement variable, indicating that as there are more negative impacts of the action, the less likely it is judged permissible to cut in line.  Universality has the strongest negative correlation with the moral judgement variable as well, indicating that participants seem to consider the question ``what if everyone did this'' (the System 2 contractualist method of reasoning) when deciding if it was OK to cut.

\section{Modelling and Reasoning with Preferences} %Preference Modelling and Reasonings in Computer Science and CP-nets}

According to Sen, moral judgements are a form of preferences, driven by moral reasoning \cite{Sen}. The issue of modelling and reasoning with preferences in an AI system has been the subject of a very active research area for many years, and has produced many frameworks to deal with preferences and embed them into an AI decision making system. Different frameworks differ on properties related, for example, to expressivity, computational complexity, and easiness of preference elicitation \cite{rossi2019preferences}. In this paper we propose a novel extsnison to CP-nets, a popular formalism for reasoning with preferences \cite{cpnets,loreggia2020modeling}. We first provide background on CP-nets and how one may use them to model the questions in our experiments.

%%%THESE SECTIONS WERE MOVED TO EARLIER IN THE PAPER
%As mentioned earlier, moral judgements are strongly related to preferences over states of the world. So, in order to embed moral judgements into an artificial agent, we exploit an existing preference modelling and reasoning framework and we generalize it to include also scenario and evaluation variables.
%
%Conditional Preference networks (CP-nets) are a graphical model for compactly representing conditional and qualitative preferences \cite{cpnets}. CP-nets are comprised of sets of {\em ceteris paribus} preference statements (cp-statements). For instance, the cp-statement, {\em ``I prefer red wine to white wine if meat is served,"} asserts that, given two meals that differ {\em only} in the kind of wine served {\em and} both containing meat, the meal with red wine is preferable to the meal with white wine.  CP-nets have been extensively used in the preference reasoning, preference learning, and social choice literature as a formalism for working with qualitative preferences \cite{DHKP11a,RVW11a,BCELP16a}. CP-nets have even been used to compose web services \cite{wang2009web} and other decision aid systems \cite{pu2011usability}. While there are many formalisms to choose from when modeling preferences, we focus on CP-nets as they are graphical and intuitive.  Transforming the results of our experiments into other formalisms is an interesting and important direction for future work.

Formally, a CP-net has a set of features (or variables) $F = \{X_1,\ldots,X_n\}$ with finite domains $ D(X_1),\ldots, D(X_n)$. For each feature $X_i$, we are given a set of {\em parent} features $Pa(X_i)$ that can affect the preferences over the values of $X_i$. This defines a {\em dependency graph} in which each node $X_i$ has $Pa(X_i)$ as its immediate predecessors. An {\em acyclic} CP-net is one in which the dependency graph is acyclic. 
Given this structural dependency information among a CP-net's variables, one needs to specify the preference over the values of each variable $X_i$ for {\em each complete assignment} to the parent variables, $Pa(X_i)$. This preference is assumed to take the form of a total or partial order over $D(X_i)$. A cp-statement for some feature $X_i$ that has parents $Pa(X_i) = \{x_1,\ldots,x_n\}$ and domain $D(X_i) = \{a_1,\ldots,a_m\}$ is a total ordering over $D(X_i)$. 
% and has general form: $x_1=v_1, x_2=v_2, \ldots,x_n=v_{n} : a_1 \succ \ldots \succ a_m$, where for each $X_i \in Pa(X_1): x_i=v_i$ is an assignment to a parent of $X_i$ with $v_i \in \cal{D} (X_i)$. 
The set of cp-statements regarding a certain variable $X_i$ is called the cp-table for $X_i$. For example, given a CP-net with features $A$, $B$, $C$, and $D$.  Each variable has binary domain containing $f$ and $\overline{f}$ if $F$ is the name of the feature. Here, statement $a \succ \overline{a}$ represents the unconditional preference for $A=a$ over $A=\overline{a}$, while statement $c: d \succ \overline{d}$ states that $D=d$ is preferred to $D=\overline{d}$, given that $C=c$.

Among the several generalizations of CP-nets, in this work we focus on an expansion of the traditional CP-nets model to model cases where individuals have indifference over the values of some features, that is, some preference information is missing \cite{allen2013cp}. %, modeled using incomparability. 
A cp-statement $z: a_i \approx a_j$ means that the person is indifferent between $a_i$ and $a_j$ given the assignment to the parents variables $z$, i.e. $a_i \succeq a_j$ and $a_j \succeq a_i$. A lack of information over one of the values of a variable is modeled with empty cp-statements for that value.

\begin{example}
\label{example1}
John is in line at the airport. When he is the first of the line, a person approaches him and kindly ask to cut the line because her flight is going to leave in 5 minutes. Due to security constraints, he always prefers not to let anyone cut the line at the airport. He has however different preferences when he is not at the airport: usually he lets a person to cut the line if he is on time while he prefers not to let someone to cut the line if he is in late with his daily schedule.
\end{example}

\begin{figure}[!t]
\centering
%\begin{minipage}[t]{0.42\textwidth}
\scriptsize
\centering
\resizebox{0.6\linewidth}{!}{
\begin{tikzpicture}
\tikzstyle{every node}=[draw,shape=circle,fill=blue!50];
\tikzset{edge/.style = {->,> = latex'}}
\node (S) at (0, 1) {$S$};
\node (T) at (0, -1) {$T$};
\node (P) at (2.5, 0) {$P$};

\draw[beats]  (S) to (P);
\draw[beats]  (T) to (P);

\coordinate [label=left: {
\begin{tabular}{c}
$a, \bar{a}$ \\
\end{tabular}
}] (p) at (0,1);

\coordinate [label=left: {
\begin{tabular}{c}
$o > \bar{o}$ \\
\end{tabular}
}] (p) at (0,-1);

\coordinate [label=right: {
\begin{tabular}{cc}
$ a,o: $	& $ \bar{c} > c$ \\
$ a, \bar{o}: $	& $\bar{c} > c$ \\
$ \bar{a}, o: $	& $ c > \bar{c}$ \\
$ \bar{a}, \bar{o}: $	& $ \bar{c} > c$ \\
\end{tabular}
}] (p) at (1.5,-.9);

\end{tikzpicture}
}
%\end{minipage}
\caption{The CP-net representing John's preferences described in Example \ref{example1}. Variable $S$ denotes the scenario: $a$ for ``at the airport'', and $\overline{a}$ for ``not at the airport''; variable $T$ represents time: $o$ for ``on time'', and $\overline{o}$ for ``not on time''; and a preference variable $P$ represent the judgement over cutting the line: i.e., $c$ is for ``ok to cut the line'', and $\overline{c}$ is for ``not ok to cut the line''.}
\label{excpnet1}
\end{figure}
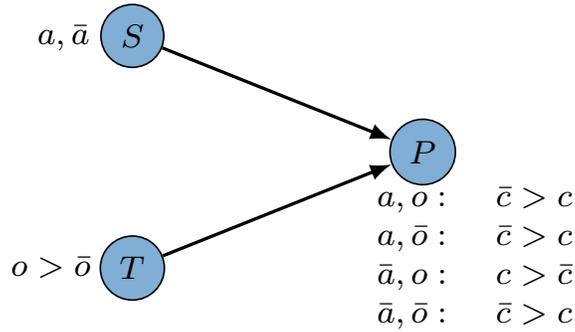

The CP-net in Figure \ref{excpnet1}, with features $S$, $T$, and $P$, represents John's preferences as described in Example \ref{example1}.  Specifically, variable $S$ denotes the scenario and takes the values $a$ for "at the airport", and $\overline{a}$ for "not at the airport". while variable $T$ represents time, and takes value $o$ for "on time" and $\overline{o}$ for "not on time".  Finally we have our preference variable $P$ which has values for cut ($c$) and not cut ($\overline{c}$). Observe that we do not report any preference between the values of variable $S$ and hence there is no cp-table for that variable. For  variable $P$, %the preference variable, 
we have a complete preference order over his domain for each possible combination of values of the parent variables $P$ and $T$.

%%=============================================================
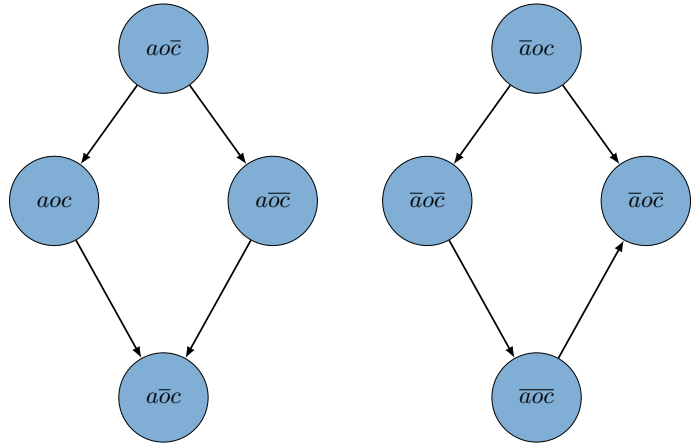
\begin{figure}
\centering
\resizebox{0.6\linewidth}{!}{%
\begin{tikzpicture}
\tikzstyle{every node}=[draw,shape=circle,fill=blue!50];
\tikzset{edge/.style = {->,> = latex'}}

%\node[label={[label distance=0.1]20:Most Preferred}] (a) at (-3, 0)
\node (a) at (-3, 0){\begin{tabular}{c}
$ao\overline{c}$ \\
\end{tabular}
};

\node[below=of a, xshift=-50] (b) 
{\begin{tabular}{c}
$aoc$ \\
\end{tabular}
};

\node[below=of a, xshift=50] (c) 
{\begin{tabular}{c}
$a\overline{oc}$ \\
\end{tabular}
};

\node[below=of a, yshift=-90] (d) {\begin{tabular}{c}
$a\overline{o}c$ \\
\end{tabular}
};

\draw[beats]   (a) -> (b);
\draw[beats]   (a) -> (c);
\draw[beats]   (b) -> (d);
\draw[beats]   (c) -> (d);

%\node[label={[label distance=0.1]20:Most Preferred}] (e) at (3, 0)
\node (e) at (3, 0){\begin{tabular}{c}
$\overline{a}oc$ \\
\end{tabular}
};

\node[below=of e, xshift=-50] (f) 
{\begin{tabular}{c}
$\overline{a}o\overline{c}$ \\
\end{tabular}
};

\node[below=of e, xshift=50] (g) 
{\begin{tabular}{c}
$\overline{a}o\overline{c}$ \\
\end{tabular}
};

\node[below=of e, yshift=-90] (h) {\begin{tabular}{c}
$\overline{aoc}$ \\
\end{tabular}
};

\draw[beats]   (e) -> (f);
\draw[beats]   (e) -> (g);
\draw[beats]   (f) -> (h);
\draw[beats]   (h) -> (g);

\end{tikzpicture}
}
%\end{minipage}
\caption{The preorder induced by the CP-net in Figure \ref{excpnet1}.  The two components are due to the indifference on the domain of the variable $A$.}
\label{excpnet2}
\end{figure}

The semantics of CP-nets depends on the notion of a {\em worsening flip}: a change in the value of a variable to a less preferred value according to the cp-statement for that variable. One outcome $\alpha$ is {\em preferred to} or \emph{dominates} another outcome $\beta$ (written $\alpha \succ \beta$) if and only if there is a chain of worsening flips from $\alpha$ to $\beta$. This definition induces a preorder (i.e. a binary relation which is reflexive and transitive) over the outcomes. Indifference induces loop between pairs of outcomes while the lack of information induces incomparability in the preorder (i.e. given two outcomes $o,p$, if neither $o \preceq p$ nor $p \preceq o$  are valid, then we say that $o$ and $p$ are incomparable, denoted with $o \bowtie p$.). This incomparability makes the preference graph disconnected. In particular, in the induced preference graph there is a connected component for each combination of values of the variables with missing cp-statements.

%%=============================================================
For instance, Figure \ref{excpnet2} gives the full induced preference order for the CP-net shown in Figure \ref{excpnet1}. The component on the left side describes preferences over the airport scenario ($S=a$), while the component on the right side describes John's preferences hen not at the airport ($S=\overline{a}$). Clearly, outcomes in different components cannot be compared because they describe preferences over different scenarios. In this, incomparability is a useful too that allows us to model different scenarios.
%%=============================================================
While CP-nets are usually a compact way to express preferences, their induced order can be exponentially larger. This is why it is important to reason on CP-nets and not on their semantics. Two useful questions that one may ask are related to optimality and dominance
%The complexity of dominance and consistency testing in CP-nets is an area of active study in preference reasoning
\shortcite{goldsmith2008computational,RVW11a}. Finding the optimal outcome of a CP-net is NP-hard~\cite{cpnets} in general but can be found in polynomial time for acyclic CP-nets, by assigning the most preferred value (according to the CP--tables) for each variable in the order given by the dependencies. %Indeed, acyclic CP-nets induce a lattice over the outcomes.% as depicted in Figure \ref{excpnet1} (right).  The induced preference ordering, Figure \ref{excpnet1} (right), can be exponentially larger than the CP-net Figure \ref{excpnet1} (left).
On the other hand, checking the dominance between two outcomes is a computationally difficult problem.

\section{Moral Judgements, Fast and Slow Thinking, and Preferences}

In a standard CP-net, there is only one kind of variable: those needed to express preferences. There is no ability to describe the context in which a preference-based decision making process takes place, nor to model other auxiliary evaluation variables that may be needed, or useful, to declare our own preferences. In some sense, a CP-net is a useful tool only when it is clear what the context is, and if no reasoning on the context is needed in order to state the preferences, or when such reasoning takes place outside the CP-net formalism.
%
%\subsection{Thinking Fast and Slow about Morality}
%

%To modify the CP-net formalism, we start by taking inspiration from the ideas of Daniel Kahneman \cite{kahneman2011thinking} and his description of the two systems, System 1 and System 2, that are relevant for human decision making. The idea is that, when making decisions, humans employ two different systems: the first is reactive and makes immediate responses that are hard and fast, but sometimes wrong.  In other cases a longer term thought process is invoked and humans think about all the factors that go into a decision before making it.  There have been proposals to extend this model into reasoning and preference systems in computer science in a principled and exact way \cite{rossi2019preferences}.

Given this limitation, there have been calls to extend this model into reasoning and preference systems in computer science in a principled and exact way \cite{rossi2019preferences}. We therefore modified the CP-net formalism to be able to capture the psychological mechanisms at play in our subjects' moral judgments.  As described in detail in Section 3, we hypothesized that subjects were using a combination of System 1 (rule-based) and System 2 (consequentialist and contractualist) thinking.  We attempt to model this two-modality reasoning process in the CP-net formalism so that we can embed it within a machine, to allow the machine to reason both fast and slow about ethical principles \shortcite{LoMaRoVe18,booch2021thinking}.  Our motivation is to extend the semantics of CP-nets in order to model both the snap judgements that do not take into account the particularities of the scenario, System 1 thinking, as well as provide the ability to reason about these details, using System 2 thinking, if necessary \cite{glazier2021making}.
  
\begin{figure*}
  \centering
  \input{sep_net}
  \caption{Our model that blends the notion of preferences with that of Scenario and Evaluation Variables.  While individuals cannot set or have preferences over the Scenario Variables, they will possess their own subjective evaluations over the Evaluation Variables given a setting to the Scenario Variables.  Given both the Scenario Variables and the Evaluation Variables, the agent can then decide on a preference over the single Preference Variable.}
  \label{fig:model}
  \end{figure*}
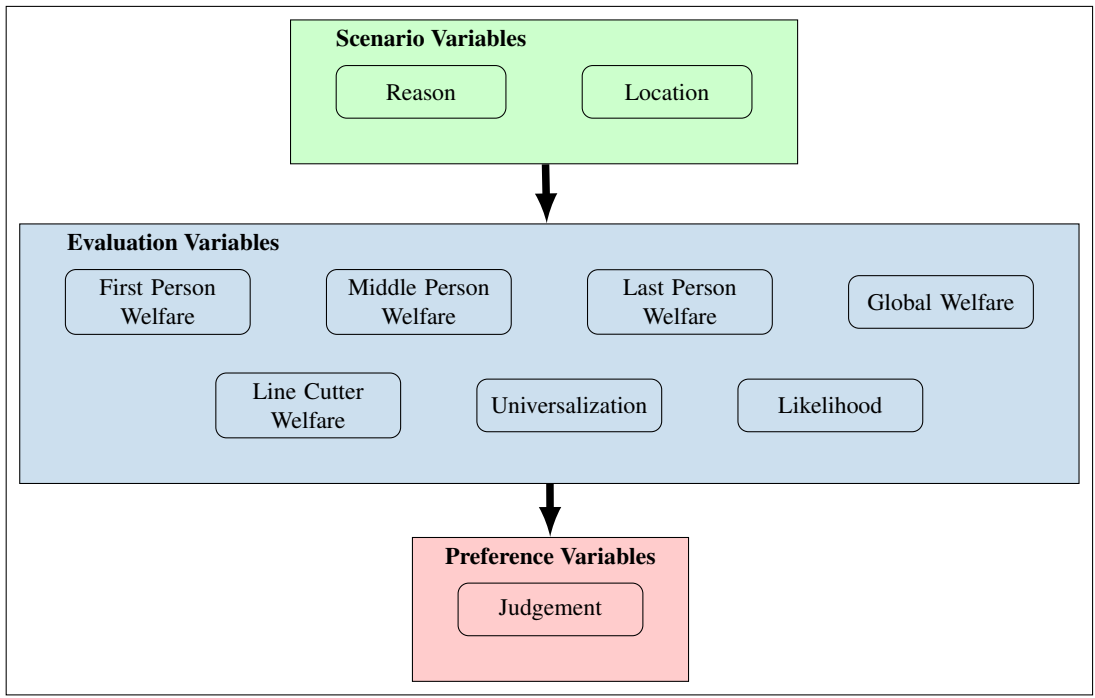

\section{SEP-nets: Scenarios, Evaluation, and Preferences}
%for Morality Driven Preferences}

Analyzing the scenarios of our vignettes and 
our conjecture of how the subjects reason about the scenarios and then respond to the single preference question (whether somebody should be allowed to cut the line or not), we propose a generalization of the CP-net formalism to handle variables associated with the context. We propose extending the formalism consisting of:
\begin{itemize}%[noitemsep]
    \item a set of \emph{scenario variables (SVs)} to define a decision making context over which there is no preference to be stated,
    \item a set of \emph{evaluation variables (EVs)} to model the evaluation process that takes place in the subjects' minds while reasoning over the given context, 
    \item to decide their preference over the \emph{preference variables (PVs)} that are already modelled in CP-nets.
\end{itemize}

%We conjecture that extending existing preference formalisms common in computer science would require new types of variables.  We have depicted this proposed layout in 
Figure \ref{fig:model} describes this generalized 
preference framework visually using all the variables from our experiments, that include the following 
%.  Starting from the CP-net formalism, we clarify the necessary 
kinds of variables:
\begin{enumerate}
\item \textbf{Scenario Variables.} A set of variables that describe the context, such as location,  whether or not the agent had already waited in line, whether or not the agent was using the main function of the line, and the size of the line.  In addition, we need a variable to specify the main reason or motivation for cutting the line. We observe that the agent does not have the ability to set values for these variables, nor does the agent have preferences over their values, as these values are set by the environment or context within which the decision is taking place. These variables do not depend on any other variable (that is, there is no incoming dependency arrow), meaning that this is part of the input to a decision making AI system.

\item \textbf{Evaluation Variables.} A set of variables that a person (or an AI system) considers (and estimates the value of) to reason about the given scenario. These are, for example, the well-being of the first in line, the well-being of the cutter, and others as discussed in the experimental details section.
%in first page of current draft).
In our experiments we have 8 of such variables.
These variables have a real valued range as a domain and the user selects one point in the range, which represents her estimate for that variable's value. However, no preferences for the values of these variables is required.
All the evaluation variables depend on the scenario variables. This follows our conjecture that people need to examine the specific scenario in order to start an evaluation phase in which they identify the evaluation variables and estimate a value for them.
%I am not sure I like the term "evaluation variable", since it may refer to evaluating anything, What about "introspection variables"?

\item \textbf{Preference Variables.} These are already included in a standard CP-net. In the setting under study, the agent expresses preference over a single value, that models whether or not, given both the values of the SVs and the values for the EVs, it is acceptable to cut the line, i.e., the moral judgement. The single preference variable depends on the evaluation variables. This again follow from the conjecture that a person needs to first perform a level of consequentialist or contractualist estimation in order to decide whether the rule, that states that a line cannot be cut, can be violated.
\end{enumerate}

As noted above, CP-nets and their variants, e.g., probabilistic CP-nets \shortcite{CGMR+13a}, 
%We can see from the above discussion that we cannot really fit the results of the experiment into the CP-net or the PCP-net formalism.  CP-nets and PCP-nets 
allow only for preference variables, and there is no option for creating a dependency between the preference variables to scenario and context variables.  We rather envision a three-layer generalization where, as shown in Figure \ref{fig:model},
the single preference variable depends on the evaluation variables, which in turn depend on the scenario variables. However, 
%At current it appears that one way is to introduce these new types of variables, as depicted in Figure \ref{} wherein all preference variables have a dependency link to a subset of evaluation variables which in turn have a dependency link to a subset of scenario variables.  In the diagram we have provided these are all interconnected but on could see, e.g., 
a finer grained analysis may show that there are evaluation variables that do not depend on the scenario variables. For example, 
the evaluation variable that has to do with the likelihood of the event happening does not have any relationship with whether or not the cutter is concerned with the main function of the line.

\subsection{SEP-nets: Formal Syntax and Semantics}

In this section, we formally define a generalization of CP-nets, called SEP-nets. This will allow us to have a compact framework to express and reason about knowledge on moral preferences. We call this generalization SEP-nets since it includes three kinds of variables, representing Scenarios, Evaluations, and Preferences. When SEP-nets include only preference features, they collapse to being standard CP-nets.

%Based on our observations and analysis in the last two sections, we generalize the model proposed by \citet{allen2013cp} to CP-nets that are capable of handling morally driven preference.  This model fits our scenario as extends the semantics of traditional CP-nets \cite{cpnets} to include indifference and incomparability. For moral CP-nets incomparability is used to model components in the preference graph whose nodes cannot be compared. For instance, outcomes over the specific scenario of the airport cannot be compared with outcomes of another scenario because the context cannot be chosen by the individual. 

\begin{definition}
An SEP-net consists of
\begin{itemize}
    \item A set of features (or variables) $V=S \cup E \cup P$. Given a variable $x$, the domain of $x$ is a finite set if $x \in S$ or $x \in P$, while it is an ordered set if $x \in E$. For simplicity, we assume 
    this ordered set is a numerical range and we define it by its minimum and maximum element (min and max). Each variable can be only in one of three sets, that is, $S \cap E= \emptyset$, $S \cap P= \emptyset$, and $P \cap E= \emptyset$.
    
    \item As in a standard CP-net, each variable $x$ has a set of parent variables $Pa(x)$, on which it depends on. However, $Pa(x) = \emptyset$ if $x \in S$; $Pa(x) \subseteq S \cup E$ if $x \in E$, and $Pa(x) \subseteq S \cup E \cup P$ if $x \in P$.
    This models a three-level acyclic structure, where scenario variables are independent, evaluation variables can depend only on scenario variables or other evaluation variables, and preference variables can depend on any variable. 
    
    \item If $x \in E$, an evaluation function $ef(x)$ identifies a single value in the domain of $x$, between min and max. Thus $ef: (min,max) \rightarrow [min,max]$. This function gives an estimate of the value for that variable.
    If $x \in P$, a standard CP-table states the preferences over the domain of $x$: for each combination of values in $Pa(x)$, the table provides a total order of $Dom(x)$.
    No CP-table or evaluation function is associated with scenario variables.
\end{itemize}

\end{definition}

Preference variables may depend directly on either all or a subset of the scenario variables. Moreover, they may depend on certain values of a scenario variable, but not others. For example, as we will see in our data analysis in Section \ref{sec:analysis}, the single preference variable is dependent on the location variable when the location is an airport or a bathroom.  
This direct dependency between preferences and scenarios models a sort of System 1 approach, where people make a moral judgement (or any other preference decision) by just looking at the situation at hand and without performing any sophisticated reasoning. On the other hand, when preference variables depend on evaluation variables, which in turn depend on scenario variables, we model a sort of System 2 approach, where people consider a scenario and a preference question, and make  an estimate of the consequences of the various options before finalizing their decision or judgement.

When there are no scenario nor evaluation variables, but just preference variables, this is a standard CP-net, so it semantics is defined as usual for CP-nets \cite{cpnets}. When instead we have a full SEP-net, its semantics is an order over all the SEP-outcomes, each being 
an assignment of values to all the variables in their respective domains. Given two outcomes 
$o = [s_1,\ldots,s_n,e_1,\ldots,e_m,p_1,\ldots,pk]$ and $o' = [s'_1,\ldots,s'_n,e'_1,\ldots,e'_m,p'_1,\ldots,p'k]$, we have $o < o'$ if 
\begin{itemize}
    \item $[s_1 \ldots, s_n] = [s'_1 \ldots, s'_n]$. This means that we are considering the same scenario.
    \item $[e_1 \ldots, e_m] = [e'_1 \ldots, e'_m] = [ef(e_1), \ldots, ef(e_m)]$. That is, the evaluation variables are set to their estimates as given by their estimation functions.
    \item $[p_1 \ldots, p_k] <_p [p'_1 \ldots, p'_k]$ in the order $<_p$ induced by the CP-net which is obtained by just considering the preference variables (in $P$) and the dependencies among them.
\end{itemize}

So, outcomes that differ on the scenario are not connected in the order induced by a SEP-net. Moreover, any outcome with the evaluation variables set to values that are different from the value given by their estimate function are not connected to any other outcome. If the dependencies among preference variables define an acyclic graph, the result is a set of partial orders, one for each scenario. Each of such partial orders has the same shape as the induced order of the CP-net obtained by the given SEP-net by setting the scenario variables to any value in their domain and the evaluation variables to their estimate value.

Given an SEP-net and its induced set of partial orders, the optimal outcomes are the top elements in the induced partial orders, thus one for each scenario, and can be efficiently computed by 1) choosing any scenario, 2) setting the evaluation variables to their estimate value, given the chosen scenario, and 3) taking the most preferred values for the preference variables.

\subsection{From Experimental Data to a SEP-net}\label{sec:analysis}

% \begin{center}
% \begin{table*}
% \caption{Average response time and standard deviation for the second dataset.}
% \begin{tabular}{ lccc } 
%  \hline
%  Context & Number of scenarios & Number of answers & Average Response Time (STD)\\
%  \hline
% Bathroom & 7 &  138 &629.2754 (372.5122)\\
% Deli & 12 & 131 & 915.8092 (504.2996)\\
% Airport & 6 & 131 & 623.0763 (414.7146)\\
%  \hline
% \end{tabular}
% \label{tab:response_time2}
% \end{table*}
% \end{center}

\begin{figure*}
  \centering
  \input{sep_emp}
  \caption{The SEP-net corresponding to the data collected in our study. SVs influence the way individuals evaluate each scenario and make a decision. For the sake of readability, we group evaluation variables based whether they depend on a particular SV in order to reduce the number of arrows. Given the SVs and the EVs, the agent can then decide on a preference over the single PV. Note that \emph{Line Cutter Welfare} does not have any effect on the PVs, while the \emph{Already Waited} variable is completely missing}
  \label{fig:fromdata}
  \end{figure*}
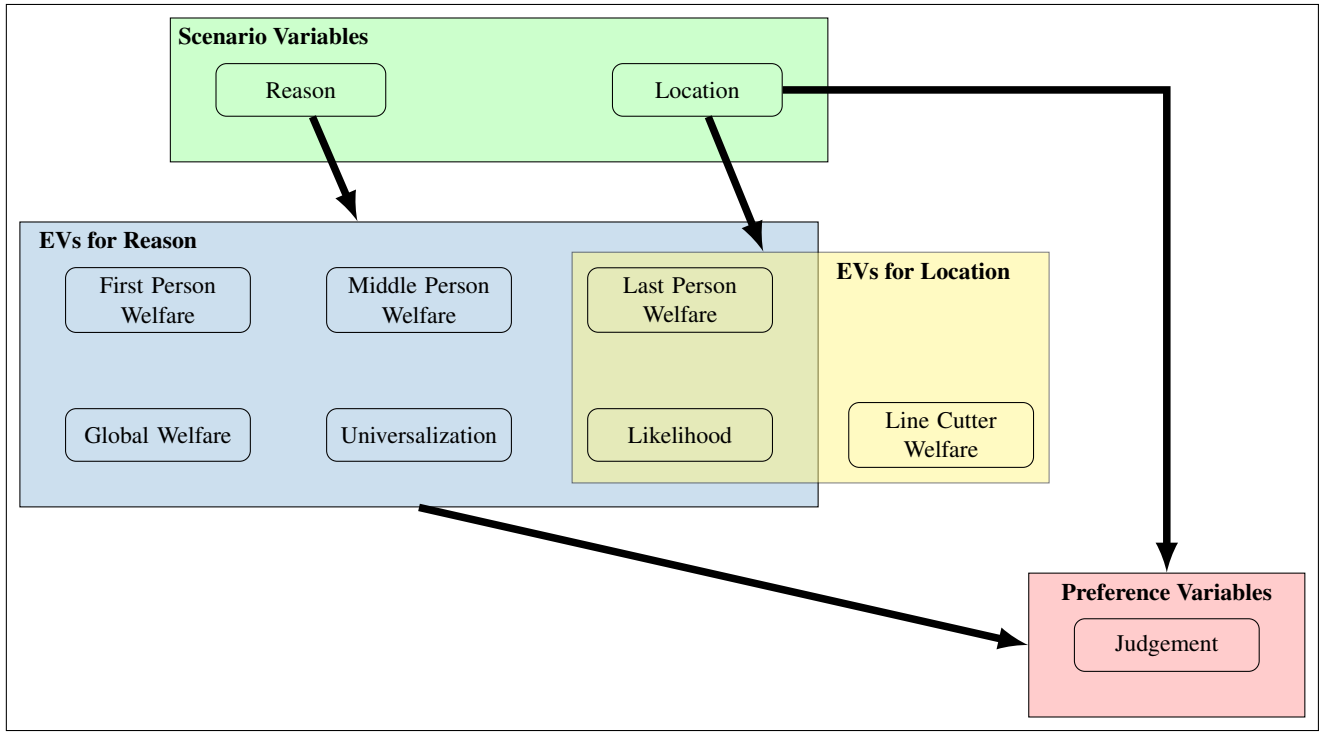

Starting from our study, we will now build a corresponding SEP-net. To do that, we must identify the variables and the dependencies among them as detailed in our vignettes from Section \ref{sec:exp_details}. The variables of this SEP-net can be seen in Figure \ref{fig:fromdata}. The scenario variables (SVs) describe all features of a scenario, and therefore refer to Reason and Location. Reason has a domain that includes all 25 reasons for cutting the line as listed in Table \ref{tab:scenarios} and Location has domain $\{Airport, Bathroom, Deli\}$. The 7 evaluation variables (EVs) in Figure \ref{fig:fromdata} correspond to the evaluation questions asked to the subjects. All have the domain $[-50,50]$, except Likelihood that has the domain $[0,100]$. There is only one preference variable (PV), corresponding to the moral judgement question. Its domain is simply yes/no.

In order to build the SEP-net corresponding to the data collected in our survey, we need to understand which SVs influence the way individuals respond to EVs. If we can find a relationship between these variables, then we say that EVs depend on SVs and validate our model. We also check whether SVs influence the PV. To test for dependency, we run Wilcoxon signed-rank tests \cite{mann1947test}, a non-parametric t-test for comparing paired data samples from the evaluations of individuals. Using this we test if the following four null hypotheses can be rejected:
\begin{enumerate}%[noitemsep]
    \item NH1: Location does not affect the EVs;
    \item NH2: Reason does not affect the EVs;
    \item NH3: Location does not affect the PV;
    \item NH4: Evaluation variables do not affect the PV.
\end{enumerate}

%We will now explain the results of our analysis and the implication for our SEP-net.

%Initially we checked whether the order of questions may change individuals' answers. We split people into two groups: one group first evaluate a scenario and then answer whether the cutter is allowed to cut the line, the other way round for the other group. The results of this analysis are reported in Table \ref{tab:anova}, as you can notice the null-hypothesis is not rejected for all the cases, thus we derive that the order does not influence the preference variable.

\paragraph{NH1: Location Does Not Affect the Evaluation Variables.}
In order to determine this, we ran a Wilcoxon signed-rank test for each pair of locations and evaluation variable.  The results of this are depicted in Table \ref{tab:nh1}.

\input{tab_nh1}

NH1 is only rejected in a few cases, specifically between Deli and Bathroom when evaluating the effect on the last person, cutter, and likelihood as well as between the Deli and Airport when evaluating the effect on the cutter and likelihood. Hence we can say that \textbf{NH1 is only partially rejected, and that some Evaluation Variables are affected by the Location, specifically: last person, cutter, and likelihood}. We have depicted these results in Figure \ref{fig:fromdata} where we build an SEP net from our experimental data. It is interesting to notice that there are a set of variables that seem to be independent of location.  This implies that no matter what location, the value of the EV does not change. Hence, it may be that some rules, no matter what is happening, are never ok to violate.

\paragraph{NH2: Reason Does Not Affect the Evaluation Variables}

For each pair of scenarios, we checked the Wilcoxon signed-rank test for each of the evaluation variables; we omit the full $7 \times {25 \choose 2}$ pairs for readability. For NH2 we can only reject the null hypothesis in some cases, however, \textbf{some evaluation variables are significantly affected for each assignment to the reason variable}.  Hence, we can say that individuals evaluate the scenario differently based on what is happening, and that the reason can and does influence all evaluation variables, with the exception of the Line Cutter variable, as shown in Figure \ref{fig:fromdata}.

\paragraph{NH3: Location Does Not Affect the PV}

We investigated whether or not the location had an effect on the preference variable. In order to test this, we selected four reasons for each location, since there were different numbers per location, aggregated these, and compared the mean responses to the to the Preference Variable or moral judgement. The complete results are depicted in Table \ref{tab:nh3}.

\input{tab_nh3}

From this we can reject NH3 for all pairs except Deli and Bathroom.  \textbf{This indicates that in some cases location may be sufficient to evaluate the vignette and make a decision.}. This is represented by the arc between location and preference variable in Figure \ref{fig:fromdata}.  It is interesting to note that participants seemed to evaluate the airport as being a significantly different location than the deli or the bathroom.

\paragraph{NH4: Evaluation Variables Do Not Affect the Preference Variable}
In order to assess the influence of evaluation variables over the preference variable we need to place them all in a comparable range. To do this we compute quartiles of each evaluation variable to bucket them and then compare the effect of this groups on the decision using the Wilcoxon signed-rank test. For each combination of Scenario and Evaluation Variable, we pair groups in order to perform the test. The full results are available in the appendix.

\textbf{The results suggest that the evaluation variables have some influence on the preference variable, except for the cutter variable.}  Indicating that in general participants were not thinking of the benefit to the cutter, but only the cost to others. This is depicted in Figure \ref{fig:fromdata} as we have moved the cutter variable out of the box of the rest of the evaluation variables, since is perceived differently. 

Based on the findings we can construct a partial graph of the dependencies between variables, the resulting SEP-net corresponding to the data of our study is shown in Figure \ref{fig:fromdata}.

\section{Conclusion and Future Work}

We have taken a first step to model and understand the question of when humans think it is morally acceptable to break rules.
%; and how, why, and when humans decide to do so. 
We constructed and studied a suite of hypothetical scenarios relating to this question, and collected human judgments about these scenarios. We showed that existing structures in the preference reasoning literature are insufficient for modeling this data, and we defined a generalization of CP-nets, called SEP-nets, which allow the linkage of preferences with scenarios and context evaluation.

Using this novel technique, we showed that humans seem to employ a complex set of preferences when determining if it is morally acceptable to break a previously established rule.  Subjects seem to take into account the particular elements of a given scenario, including location and reasons to break the rule.  Moreover, we found evidence that System 2 processes were operative in rule-breaking decisions: moral judgments were influenced by calculations of the impact of rule-violations on the well-being of others (consequentialist reasoning) as well as notions of what everyone would agree to (contractualist reasoning).  Other times, System 1 reasoning seemed to be operative; sometimes, outcomes and agreement had no relationship to moral permissibility and rather rules were considered inviolable.  Together, this pattern of data begins to suggest that moral rigidity and moral flexibility are driven by fast and slow thinking.

%We look towards extending this into other established areas of AI research and other preference formalisms.  
%
There are several ways in which we plan to extend this work. We mention here two concrete directions:

\paragraph{Comparing Preferences and Measuring Value Deviation.}
%Generalizing CP-nets to Model Moral Preferences and Other Generalizations.} 
We defined generalized CP-nets (called SEP-nets) in a way that is consistent with classical CP-nets and probabilistic CP-nets. The aim is to understand how to use a (generalized) preference structure to effectively learn and reason with morality-driven preferences, and to embed them into an AI system.  Another fruitful direction for future work is to attempt to adapt our results to other preference reasoning formalisms, e.g., soft-constraints or weighted logical representations \cite{rossi2019preferences}.  Reasoning about preferences and decision making requires also being able to compare preferences. For instance, in a multi-agent system, it is important to understand whether an agent is deviating from a societal priority or norm. To do that, recent studies proposed metric spaces over preferences \cite{loreggia2018distance} which can be used to implement value alignment procedures. Such metric spaces may be extended to be applicable to SEP-nets.  Such metrics would be useful to allow a system designer to intervene in cases where individuals (or artificial agents) behave differently than expected, or are operating outside the norms or rules of a society.

\paragraph{Prescriptive Plans Based on Moral Preferences.}
The AI research community has not only been active in understanding how to make single decisions based on preferences, but also on creating plans, consisting of sequences of actions, that would respect or follow certain preferences ~\cite{brafman2005planning}. This work can be exploited to extend the use of the moral preferences discussed in this paper into more prescriptive AI techniques such as automated planning~\cite{benton2012temporal}. Although prior efforts from the planning perspective all investigate the generation of plans that take into account pre-specified utilitarian preferences, the question of where those utilities and preferences manifest from has not been addressed very adequately so far. We are currently actively investigating methods that seek to use the data collected in this work to automatically generate preferences in the notation used by planning formalisms~\cite{gerevini2006plan}. The generation of such preferences will in turn enable us to generate prescriptive plans for agents or systems that conform to the moral standards of that agent or system. Specifically, we will transform the problem from a classification-based setting into a generative model, and then present plan (action) alternatives that agents can choose from. To situate this in the context of the current question of study: this extension would enable us to move from determining whether it was acceptable to break a rule, to generating ways to do so that are most in accordance with some preference and cost function that takes moral obligations into account.

\bibliographystyle{theapa}
\bibliography{biblio}

\clearpage
\onecolumn
\appendix

\section{Complete Statistical Tables}
\input{tab_anova}
\input{tab_nh4}
\input{tab_nh2}

\end{document}

%% file: sep_net.tex
\begin{tikzpicture}[show background rectangle, font=\footnotesize]
    % Scenario Variables
    \node[module, text width=2cm] (reason) {Reason};
    \node[module, right=of reason, text width=2cm] (location) {Location};
    %\node[module, right=of location, text width=2cm] (linesize) {Already Waited};
    
    % Evaluation Variables
    \node[module, below=of reason, yshift=-1.0cm, xshift=-3
    .5cm, text width=2.2cm] (first) {First Person Welfare};
    \node[module, right=of first, text width=2.2cm] (middle) {Middle Person Welfare};
    \node[module, right=of middle, text width=2.2cm] (last) {Last Person Welfare};
    \node[module, right=of last, text width=2.2cm] (global) {Global Welfare};
    
    \node[module, below=of first, xshift=2cm, yshift=0.5cm, text width=2.2cm] (cutter) {Line Cutter Welfare};
    \node[module, right=of cutter, text width=2.2cm] (univ) {Universalization};
    \node[module, right=of univ, text width=2.2cm] (likelihood) {Likelihood};
    
    %% Preference Variables
    \node[module, below=of univ, yshift=-1.0cm, xshift=-0.25cm, text width=2.2cm] (judge) {Judgement};
    
    % Background Grouping.
    \begin{pgfonlayer}{background}

        \node[fit=(reason) (location), draw, fill=green!20, inner sep=6mm, label={[xshift=-15mm,yshift=-5mm] \textbf{Scenario Variables}}] (SV) {};
        
        \node[fit=(first) (middle) (last) (global) (cutter) (univ) (likelihood), draw, fill=blue!20, inner sep=6mm, label={[xshift=-50mm,yshift=-5mm] \textbf{Evaluation Variables}}] (EV) {};
        
        \node[fit=(judge), draw, fill=red!20, inner sep=6mm, label={[xshift=0mm,yshift=-5mm] \textbf{Preference Variables}}] (PV) {};
    
    \end{pgfonlayer}
    
    \draw[beats, line width=1.0mm] (SV) to (EV);
    \draw[beats, line width=1.0mm] (EV) to (PV);

\end{tikzpicture}

%% file: tab_nh1.tex
\begin{center}
\begin{table}[h!]
\centering
\begin{tabular}{ lccc } 
 \hline
 Scenario & Deli-Bath & Deli-Airpt & Airpt-Bath\\
 \hline
Global Welfare &0.2782 &0.8954 &0.3028 \\
First Person Welfare &0.1779 &0.0932 &0.9696 \\
Middle Person Welfare&0.1012 &0.1390 &0.3478 \\
Last Person Welfare&\textbf{0.0064} &0.1444 &0.2763 \\
Line Cutter Welfare&\textbf{0.0069} &\textbf{0.0123} &0.3467 \\
Universalization &0.4848 &0.4356 &0.1567 \\
Likelihood &\textbf{0.0008} &\textbf{0.0138} &0.2430 \\
%delay\_min &0.9226 &\textbf{0.0002} &\textbf{0.0018} \\
%delay\_sec &0.6702 &0.2627 &0.5309 \\
 \hline
\end{tabular}
\caption{p-values for the Wilcoxon signed-rank test against the null-hypothesis NH1:  the location does not affect evaluation variables. The null-hypothesis is rejected if $p < 0.05$, these cases are reported in bold.}
\label{tab:nh1}
\end{table}
\end{center}

%% file: tab_nh3.tex
\begin{center}
\begin{table}[h!]
\centering
\begin{tabular}{ lcc } 
 \hline
 Scenario & p-value & Rejected\\
 \hline
Deli-Bath 	 & 8.759E-01 	 & False\\
Deli-Air 	 & 1.548E-08 	 & True\\
Air-Bath 	 & 2.662E-10 	 & True\\
 \hline
\end{tabular}
\caption{p-values for the Wilcoxon signed-rank test against the null-hypothesis NH3: location does not affect the preference variable. The null-hypothesis is rejected if $p < 0.05$.}
\label{tab:nh3}
\end{table}
\end{center}

%% file: sep_emp.tex
\begin{tikzpicture}[show background rectangle, font=\footnotesize]
    % Scenario Variables
    \node[module, text width=2cm] (reason) {Reason};
    \node[module, right=of reason, xshift=2cm, text width=2cm] (location) {Location};
    
    % Evaluation Variables
    \node[module, below=of reason, yshift=-1.0cm, xshift=-1.9cm, text width=2.2cm] (first) {First Person Welfare};
    \node[module, right=of first, text width=2.2cm] (middle) {Middle Person Welfare};
    \node[module, below=of first, text width=2.2cm] (global) {Global Welfare};
    \node[module, right=of global, text width=2.2cm] (univ) {Universalization};
    
    %% Both
    \node[module, right=of middle, text width=2.2cm] (last) {Last Person Welfare};
    \node[module, right=of univ, text width=2.2cm] (likelihood) {Likelihood};
    
    %% Solo
    \node[module, right=of likelihood, text width=2.2cm] (cutter) {Line Cutter Welfare};

    %% Preference Variables
    \node[module, below=of cutter, yshift=-1.0cm, xshift=3cm, text width=2.2cm] (judge) {Judgement};
    
    % Background Grouping.
    \begin{pgfonlayer}{background}

        \node[fit=(reason) (location), draw, fill=green!20, inner sep=6mm, label={[xshift=-30mm,yshift=-5mm] \textbf{Scenario Variables}}] (SV) {};
        
        \node[fit=(first) (middle) (last) (global) (univ) (likelihood), draw, fill=blue!20, inner sep=6mm, label={[xshift=-40mm,yshift=-5mm] \textbf{EVs for Reason}}] (REV) {};
        
        \node[fit=(last) (likelihood) (cutter), draw, fill=yellow!60, opacity=.5, inner sep=2mm, label={[xshift=15mm,yshift=-5mm] \textbf{EVs for Location}}] (LEV) {};
        
        \node[fit=(judge), draw, fill=red!20, inner sep=6mm, label={[xshift=0mm,yshift=-5mm] \textbf{Preference Variables}}] (PV) {};
    
    \end{pgfonlayer}
    
    \draw[beats, line width=1.0mm] (reason) to (REV);
    \draw[beats, line width=1.0mm] (REV.south) to (PV.west);
    \draw[beats, line width=1.0mm] (location) to (LEV);
    
    \draw[beats, line width=1.0mm] (location.east) -| (PV);

\end{tikzpicture}

%% file: tab_anova.tex
\begin{center}
\begin{table*}[h!]
\begin{tabular}{ lccccc } 
 \hline
 Scenario & PREFthenEVAL & PREFthenEVAL & p-value & Rejected\\
 \hline
 DFO\_SP & 0.1212 (0.3264) & 0.1692 (0.3750) & 0.5059 & False\\
 DFO\_WA & 0.4697 (0.4991) & 0.4154 (0.4928) & 0.5860 & False\\
 DFO\_SO & 0.6970 (0.4596) & 0.6923 (0.4615) & 0.8535 & False\\
 DFO\_CA & 0.7576 (0.4285) & 0.6154 (0.4865) & 0.0743 & False\\
 DFO\_CO & 0.7424 (0.4373) & 0.6308 (0.4826) & 0.1979 & False\\
 DFO\_SU & 0.0909 (0.2875) & 0.1385 (0.3454) & 0.4281 & False\\
 DFO\_BR & 0.0758 (0.2646) & 0.1692 (0.3750) & 0.1414 & False\\
 DFO\_HS & 0.1061 (0.3079) & 0.1385 (0.3454) & 0.6379 & False\\
 DFO\_TP & 0.2121 (0.4088) & 0.1692 (0.3750) & 0.4579 & False\\
 DFO\_MA & 0.2879 (0.4528) & 0.3231 (0.4677) & 0.5860 & False\\
 DFO\_FA & 0.2424 (0.4285) & 0.3231 (0.4677) & 0.3596 & False\\
 DFO\_SW & 0.8182 (0.3857) & 0.7538 (0.4308) & 0.4236 & False\\
 BFO\_HA & 0.5507 (0.4974) & 0.5217 (0.4995) & 0.7317 & False\\
 BFO\_CL & 0.1594 (0.3661) & 0.2899 (0.4537) & 0.0971 & False\\
 BFO\_TU & 0.1014 (0.3019) & 0.1884 (0.3910) & 0.1414 & False\\
 BFO\_JA & 0.1594 (0.3661) & 0.2319 (0.4220) & 0.3248 & False\\
 BFO\_FR & 0.7101 (0.4537) & 0.7536 (0.4309) & 0.5730 & False\\
 BFO\_EL & 0.1014 (0.3019) & 0.2029 (0.4022) & 0.0948 & False\\
 BFO\_BR & 0.7391 (0.4391) & 0.7681 (0.4220) & 0.7238 & False\\
 AFO\_MN & 0.4030 (0.4905) & 0.4219 (0.4939) & 0.6096 & False\\
 AFO\_BA & 0.4328 (0.4955) & 0.4531 (0.4978) & 0.7526 & False\\
 AFO\_JA & 0.4478 (0.4973) & 0.5625 (0.4961) & 0.1088 & False\\
 AFO\_CA & 0.5224 (0.4995) & 0.5781 (0.4939) & 0.3471 & False\\
 AFO\_BR & 0.4478 (0.4973) & 0.4062 (0.4911) & 0.4285 & False\\
 AFO\_HR & 0.8209 (0.3834) & 0.7969 (0.4023) & 0.5607 & False\\
 \hline
\end{tabular}
\caption{p-values for the Wilcoxon signed-rank test against the null-hypothesis: changing the order of evaluation or preference does not influence the preference value. The null-hypothesis is rejected if $p < 0.05$.}
\label{tab:anova}
\end{table*}
\end{center}

%% file: tab_nh4.tex
{\small
\begin{center}
\begin{longtable}{llcccccc} 
%\begin{tabular}{ llcccccc } 
\hline
Scenario & EV & Group 0-1 & Group 0-2 & Group 0-3 & Group 1-2 & Group 1-3 & Group 2-3\\
\hline
\endhead
DFO\_SP & everyone &0.1489 &0.4237 &0.4840 &1.0000 &0.7656 &0.7656 \\
DFO\_WA & everyone &0.3506 &0.3458 &\textbf{0.0004} &1.0000 &\textbf{0.0050} &0.3458 \\
DFO\_SO & everyone &1.0000 &0.3458 &\textbf{0.0001} &0.7728 &\textbf{0.0001} &\textbf{0.0369} \\
DFO\_CA & everyone &0.1817 &0.0726 &\textbf{0.0000} &0.1817 &\textbf{0.0001} &0.1096 \\
DFO\_CO & everyone &0.3506 &0.0719 &\textbf{0.0000} &0.0719 &\textbf{0.0000} &nan \\
DFO\_SU & everyone &0.2986 &0.7728 &\textbf{0.0369} &1.0000 &0.4237 &1.0000 \\
DFO\_BR & everyone &0.2330 &1.0000 &1.0000 &1.0000 &0.7897 &0.7656 \\
DFO\_HS & everyone &0.1489 &1.0000 &0.7728 &0.1489 &0.7656 &0.0719 \\
\end{longtable}
\end{center}
}

%% file: tab_nh2.tex
{\small
\begin{center}
\begin{longtable}{llccccccccc}
%\begin{tabular}{ llccccccccc } 
Scenario1 & Scenario2 & everyone & first\_person & middle\_person & last\_person & cutter & univers. & likelihood  \\
\hline
\endhead
DFO\_SP & DFO\_WA & 0.2576 & 0.2927 & 0.0810 & 0.1037 & 0.5957 & \textbf{0.0393} & 0.9389 & \\
DFO\_SP & DFO\_SO & \textbf{0.0098} & \textbf{0.0238} & \textbf{0.0050} & \textbf{0.0029} & 0.1911 & \textbf{0.0007} & 0.4266 & \\
DFO\_SP & DFO\_CA & \textbf{0.0001} & \textbf{0.0005} & \textbf{0.0000} & \textbf{0.0000} & 0.1741 & \textbf{0.0000} & \textbf{0.0000} & \\
DFO\_SP & DFO\_CO & \textbf{0.0026} & \textbf{0.0117} & \textbf{0.0023} & \textbf{0.0020} & \textbf{0.0083} & \textbf{0.0033} & \textbf{0.0000} & \\
DFO\_SP & DFO\_SU & 0.1426 & 0.1960 & 0.1692 & \textbf{0.0200} & \textbf{0.0000} & 0.1937 & \textbf{0.0000} & \\
DFO\_SP & DFO\_BR & 0.1524 & 0.0510 & 0.0742 & 0.0533 & 0.6433 & 0.6679 & \textbf{0.0000} & \\
DFO\_SP & DFO\_HS & 0.6004 & 0.9483 & 0.9699 & 0.6789 & \textbf{0.0006} & 0.6405 & \textbf{0.0120} & \\
DFO\_SP & DFO\_TP & 0.1848 & 0.3437 & 0.2102 & 0.0895 & 0.8971 & 0.6394 & \textbf{0.0072} & \\
\end{longtable}
\end{center}
}

%% file: main.bbl
\begin{thebibliography}{}

\bibitem[\protect\BCAY{Alexander}{Alexander}{2012}]{alexander2012experimental}
Alexander, J. \BBOP2012\BBCP.
\newblock {\Bem Experimental philosophy: An introduction}.
\newblock Polity.

\bibitem[\protect\BCAY{Alkoby, Rath,\ \BBA\ Stone}{Alkoby
  et~al.}{2019}]{alkoby2019teaching}
Alkoby, S., Rath, A., \BBA\ Stone, P. \BBOP2019\BBCP.
\newblock \BBOQ Teaching social behavior through human reinforcement for ad hoc
  teamwork-the {STAR} framework\BBCQ\
\newblock In {\Bem Proc. of the~18th~ AAMAS}.

\bibitem[\protect\BCAY{Allen, Smit,\ \BBA\ Wallach}{Allen
  et~al.}{2005}]{allen2005artificial}
Allen, C., Smit, I., \BBA\ Wallach, W. \BBOP2005\BBCP.
\newblock \BBOQ Artificial morality: Top-down, bottom-up, and hybrid
  approaches\BBCQ\
\newblock {\Bem Ethics and Information Technology}, {\Bem 7\/}(3), 149--155.

\bibitem[\protect\BCAY{Allen}{Allen}{2013}]{allen2013cp}
Allen, T.~E. \BBOP2013\BBCP.
\newblock \BBOQ {CP}-nets with indifference\BBCQ\
\newblock In {\Bem 2013 51st Annual Allerton Conference on Communication,
  Control, and Computing (Allerton)}, \BPGS\ 1488--1495. IEEE.

\bibitem[\protect\BCAY{Arnold, Thomas, Kasenberg,\ \BBA\ Scheutzs}{Arnold
  et~al.}{2017}]{conf/aaai/ArnoldKS17}
Arnold, T., Thomas, Kasenberg, D., \BBA\ Scheutzs, M. \BBOP2017\BBCP.
\newblock \BBOQ Value alignment or misalignment - what will keep systems
  accountable?\BBCQ\
\newblock In {\Bem AI, Ethics, and Society, Papers from the 2017 {AAAI}
  Workshop}.

\bibitem[\protect\BCAY{Awad, Dsouza, Kim, Schulz, Henrich, Shariff, Bonnefon,\
  \BBA\ Rahwan}{Awad et~al.}{2018}]{awad2018moral}
Awad, E., Dsouza, S., Kim, R., Schulz, J., Henrich, J., Shariff, A., Bonnefon,
  J.-F., \BBA\ Rahwan, I. \BBOP2018\BBCP.
\newblock \BBOQ The moral machine experiment\BBCQ\
\newblock {\Bem Nature}, {\Bem 563\/}(7729), 59.

\bibitem[\protect\BCAY{Balakrishnan, Bouneffouf, Mattei,\ \BBA\
  Rossi}{Balakrishnan et~al.}{2019}]{balakrishnan2018incorporating}
Balakrishnan, A., Bouneffouf, D., Mattei, N., \BBA\ Rossi, F. \BBOP2019\BBCP.
\newblock \BBOQ Incorporating behavioral constraints in online {AI}
  systems\BBCQ\
\newblock In {\Bem Proc. of the~33rd~AAAI}.

\bibitem[\protect\BCAY{Baumard, Andr{\'e},\ \BBA\ Sperber}{Baumard
  et~al.}{2013}]{baumard2013mutualistic}
Baumard, N., Andr{\'e}, J.-B., \BBA\ Sperber, D. \BBOP2013\BBCP.
\newblock \BBOQ A mutualistic approach to morality: The evolution of fairness
  by partner choice\BBCQ\
\newblock {\Bem Behavioral and Brain Sciences}, {\Bem 36\/}(1), 59--78.

\bibitem[\protect\BCAY{Benton, Coles,\ \BBA\ Coles}{Benton
  et~al.}{2012}]{benton2012temporal}
Benton, J., Coles, A., \BBA\ Coles, A. \BBOP2012\BBCP.
\newblock \BBOQ Temporal planning with preferences and time-dependent
  continuous costs\BBCQ\
\newblock In {\Bem Proc. 22nd ICAPS}.

\bibitem[\protect\BCAY{Bonnefon, Shariff,\ \BBA\ Rahwan}{Bonnefon
  et~al.}{2020}]{bonnefon2020moral}
Bonnefon, J.-F., Shariff, A., \BBA\ Rahwan, I. \BBOP2020\BBCP.
\newblock {\Bem The moral psychology of {AI} and the ethical opt-out problem}.
\newblock Oxford University Press, Oxford, UK.

\bibitem[\protect\BCAY{Booch, Fabiano, Horesh, Kate, Lenchner, Linck, Loreggia,
  Murgesan, Mattei, Rossi, et~al.}{Booch et~al.}{2021}]{booch2021thinking}
Booch, G., Fabiano, F., Horesh, L., Kate, K., Lenchner, J., Linck, N.,
  Loreggia, A., Murgesan, K., Mattei, N., Rossi, F., et~al. \BBOP2021\BBCP.
\newblock \BBOQ Thinking fast and slow in {AI}\BBCQ\
\newblock In {\Bem Proceedings of the AAAI Conference on Artificial
  Intelligence}, \lowercase{\BVOL}~35, \BPGS\ 15042--15046.

\bibitem[\protect\BCAY{Boutilier, Brafman, Domshlak, Hoos,\ \BBA\
  Poole}{Boutilier et~al.}{2004}]{cpnets}
Boutilier, C., Brafman, R., Domshlak, C., Hoos, H., \BBA\ Poole, D.
  \BBOP2004\BBCP.
\newblock \BBOQ {CP-nets}: A tool for representing and reasoning with
  conditional ceteris paribus preference statements\BBCQ\
\newblock {\Bem Journal of Artificial Intelligence Research}, {\Bem 21},
  135--191.

\bibitem[\protect\BCAY{Brafman\ \BBA\ Chernyavsky}{Brafman\ \BBA\
  Chernyavsky}{2005}]{brafman2005planning}
Brafman, R.~I.\BBACOMMA\  \BBA\ Chernyavsky, Y. \BBOP2005\BBCP.
\newblock \BBOQ Planning with goal preferences and constraints.\BBCQ\
\newblock In {\Bem ICAPS}, \BPGS\ 182--191.

\bibitem[\protect\BCAY{Brandt, Conitzer, Endriss, Lang,\ \BBA\
  Procaccia}{Brandt et~al.}{2016}]{BCELP16a}
Brandt, F., Conitzer, V., Endriss, U., Lang, J., \BBA\ Procaccia, A.~D.\BEDS.
  \BBOP2016\BBCP.
\newblock {\Bem Handbook of Computational Social Choice}.
\newblock Cambridge University Press.

\bibitem[\protect\BCAY{Cornelio, Goldsmith, Mattei, Rossi,\ \BBA\
  Venable}{Cornelio et~al.}{2013}]{CGMR+13a}
Cornelio, C., Goldsmith, J., Mattei, N., Rossi, F., \BBA\ Venable, K.
  \BBOP2013\BBCP.
\newblock \BBOQ Updates and uncertainty in {CP}-nets\BBCQ\
\newblock In {\Bem Proc. of the~26th~AUSAI}.

\bibitem[\protect\BCAY{Cushman}{Cushman}{2013}]{cushman2013action}
Cushman, F. \BBOP2013\BBCP.
\newblock \BBOQ Action, outcome, and value: A dual-system framework for
  morality\BBCQ\
\newblock {\Bem Personality and social psychology review}, {\Bem 17\/}(3),
  273--292.

\bibitem[\protect\BCAY{Domshlak, H{\"u}llermeier, Kaci,\ \BBA\ Prade}{Domshlak
  et~al.}{2011}]{DHKP11a}
Domshlak, C., H{\"u}llermeier, E., Kaci, S., \BBA\ Prade, H. \BBOP2011\BBCP.
\newblock \BBOQ Preferences in {AI}: An overview\BBCQ\
\newblock {\Bem Artificial Intelligence}, {\Bem 175\/}(7), 1037--1052.

\bibitem[\protect\BCAY{Doris, Group, et~al.}{Doris
  et~al.}{2010}]{doris2010moral}
Doris, J.~M., Group, M. P.~R., et~al. \BBOP2010\BBCP.
\newblock {\Bem The moral psychology handbook}.
\newblock OUP Oxford.

\bibitem[\protect\BCAY{Engelmann\ \BBA\ Waldmann}{Engelmann\ \BBA\
  Waldmann}{2022}]{engelmann2022weigh}
Engelmann, N.\BBACOMMA\  \BBA\ Waldmann, M.~R. \BBOP2022\BBCP.
\newblock \BBOQ How to weigh lives. a computational model of moral judgment in
  multiple-outcome structures\BBCQ\
\newblock {\Bem Cognition}, {\Bem 218}, 104910.

\bibitem[\protect\BCAY{Gauthier}{Gauthier}{1986}]{gauthier1986morals}
Gauthier, D. \BBOP1986\BBCP.
\newblock {\Bem Morals by agreement}.
\newblock Oxford University Press on Demand.

\bibitem[\protect\BCAY{Gerevini\ \BBA\ Long}{Gerevini\ \BBA\
  Long}{2005}]{gerevini2006plan}
Gerevini, A.\BBACOMMA\  \BBA\ Long, D. \BBOP2005\BBCP.
\newblock \BBOQ Plan constraints and preferences in pddl3\BBCQ.

\bibitem[\protect\BCAY{Glazier, Loreggia, Mattei, Rahgooy, Rossi,\ \BBA\
  Venable}{Glazier et~al.}{2021}]{glazier2021making}
Glazier, A., Loreggia, A., Mattei, N., Rahgooy, T., Rossi, F., \BBA\ Venable,
  K.~B. \BBOP2021\BBCP.
\newblock \BBOQ Making human-like trade-offs in constrained environments by
  learning from demonstrations\BBCQ\
\newblock {\Bem arXiv preprint arXiv:2109.11018}.

\bibitem[\protect\BCAY{Goldsmith, Lang, Truszczy{\'n}ski,\ \BBA\
  Wilson}{Goldsmith et~al.}{2008}]{goldsmith2008computational}
Goldsmith, J., Lang, J., Truszczy{\'n}ski, M., \BBA\ Wilson, N. \BBOP2008\BBCP.
\newblock \BBOQ The computational complexity of dominance and consistency in
  {CP}-nets\BBCQ\
\newblock {\Bem Journal of Artificial Intelligence Research}, {\Bem 33\/}(1),
  403--432.

\bibitem[\protect\BCAY{Greene}{Greene}{2014}]{greene2014moral}
Greene, J.~D. \BBOP2014\BBCP.
\newblock {\Bem Moral tribes: Emotion, reason, and the gap between us and
  them}.
\newblock Penguin.

\bibitem[\protect\BCAY{Habermas}{Habermas}{1990}]{habermas1990moral}
Habermas, J. \BBOP1990\BBCP.
\newblock {\Bem Moral consciousness and communicative action}.
\newblock MIT press.

\bibitem[\protect\BCAY{Haidt}{Haidt}{2007}]{haidt2007new}
Haidt, J. \BBOP2007\BBCP.
\newblock \BBOQ The new synthesis in moral psychology\BBCQ\
\newblock {\Bem science}, {\Bem 316\/}(5827), 998--1002.

\bibitem[\protect\BCAY{Hare}{Hare}{1981}]{hare1981moral}
Hare, R.~M. \BBOP1981\BBCP.
\newblock {\Bem Moral thinking: Its levels, method, and point}.
\newblock Oxford: Clarendon Press; New York: Oxford University Press.

\bibitem[\protect\BCAY{Harsanyi}{Harsanyi}{1977}]{harsanyi1977morality}
Harsanyi, J.~C. \BBOP1977\BBCP.
\newblock \BBOQ Morality and the theory of rational behavior\BBCQ\
\newblock {\Bem Social Research}, {\Bem 44\/}(4), 623.

\bibitem[\protect\BCAY{Holyoak\ \BBA\ Powell}{Holyoak\ \BBA\
  Powell}{2016}]{holyoak2016deontological}
Holyoak, K.~J.\BBACOMMA\  \BBA\ Powell, D. \BBOP2016\BBCP.
\newblock \BBOQ Deontological coherence: A framework for commonsense moral
  reasoning.\BBCQ\
\newblock {\Bem Psychological Bulletin}, {\Bem 142\/}(11), 1179.

\bibitem[\protect\BCAY{Jiang, Hwang, Bhagavatula, Bras, Forbes, Borchardt,
  Liang, Etzioni, Sap,\ \BBA\ Choi}{Jiang et~al.}{2021}]{jiang2021delphi}
Jiang, L., Hwang, J.~D., Bhagavatula, C., Bras, R.~L., Forbes, M., Borchardt,
  J., Liang, J., Etzioni, O., Sap, M., \BBA\ Choi, Y. \BBOP2021\BBCP.
\newblock \BBOQ Delphi: Towards machine ethics and norms\BBCQ\
\newblock {\Bem arXiv preprint arXiv:2110.07574}.

\bibitem[\protect\BCAY{Kahneman}{Kahneman}{2011}]{kahneman2011thinking}
Kahneman, D. \BBOP2011\BBCP.
\newblock {\Bem Thinking, Fast and Slow}.
\newblock Macmillan.

\bibitem[\protect\BCAY{Kim, Kleiman-Weiner, Abeliuk, Awad, Dsouza, Tenenbaum,\
  \BBA\ Rahwan}{Kim et~al.}{2018}]{kim2018computational}
Kim, R., Kleiman-Weiner, M., Abeliuk, A., Awad, E., Dsouza, S., Tenenbaum,
  J.~B., \BBA\ Rahwan, I. \BBOP2018\BBCP.
\newblock \BBOQ A computational model of commonsense moral decision
  making\BBCQ\
\newblock In {\Bem Proceedings of the 2018 AAAI/ACM Conference on AI, Ethics,
  and Society}, \BPGS\ 197--203.

\bibitem[\protect\BCAY{Kleiman-Weiner, Gerstenberg, Levine,\ \BBA\
  Tenenbaum}{Kleiman-Weiner et~al.}{2015}]{kleiman2015inference}
Kleiman-Weiner, M., Gerstenberg, T., Levine, S., \BBA\ Tenenbaum, J.~B.
  \BBOP2015\BBCP.
\newblock \BBOQ Inference of intention and permissibility in moral decision
  making.\BBCQ\
\newblock In {\Bem CogSci}. Citeseer.

\bibitem[\protect\BCAY{Kleiman-Weiner, Saxe,\ \BBA\ Tenenbaum}{Kleiman-Weiner
  et~al.}{2017}]{kleiman2017learning}
Kleiman-Weiner, M., Saxe, R., \BBA\ Tenenbaum, J.~B. \BBOP2017\BBCP.
\newblock \BBOQ Learning a commonsense moral theory\BBCQ\
\newblock {\Bem Cognition}, {\Bem 167}, 107--123.

\bibitem[\protect\BCAY{Knobe}{Knobe}{2007}]{knobe2007experimental}
Knobe, J. \BBOP2007\BBCP.
\newblock \BBOQ Experimental philosophy\BBCQ\
\newblock {\Bem Philosophy Compass}, {\Bem 2\/}(1), 81--92.

\bibitem[\protect\BCAY{Levine, Kleiman-Weiner, Chater, Cushman,\ \BBA\
  Tenenbaum}{Levine et~al.}{2018}]{levine2018contractualism}
Levine, S., Kleiman-Weiner, M., Chater, N., Cushman, F., \BBA\ Tenenbaum, J.~B.
  \BBOP2018\BBCP.
\newblock \BBOQ The cognitive mechanisms of contractualist moral
  decision-making.\BBCQ\
\newblock In {\Bem CogSci}. Citeseer.

\bibitem[\protect\BCAY{Levine, Kleiman-Weiner, Schulz, Tenenbaum,\ \BBA\
  Cushman}{Levine et~al.}{2020}]{levine2020logic}
Levine, S., Kleiman-Weiner, M., Schulz, L., Tenenbaum, J., \BBA\ Cushman, F.
  \BBOP2020\BBCP.
\newblock \BBOQ The logic of universalization guides moral judgment\BBCQ\
\newblock {\Bem Proceedings of the National Academy of Sciences}, {\Bem
  117\/}(42), 26158--26169.

\bibitem[\protect\BCAY{Levine\ \BBA\ Leslie}{Levine\ \BBA\
  Leslie}{2021}]{levine-preschool-trolley}
Levine, S.\BBACOMMA\  \BBA\ Leslie, A. \BBOP2021\BBCP.
\newblock \BBOQ Preschoolers use the means principle to make moral
  judgments\BBCQ\
\newblock {\Bem psyarxiv.com/np9a5}.

\bibitem[\protect\BCAY{Lieder\ \BBA\ Griffiths}{Lieder\ \BBA\
  Griffiths}{2020}]{lieder2020resource}
Lieder, F.\BBACOMMA\  \BBA\ Griffiths, T.~L. \BBOP2020\BBCP.
\newblock \BBOQ Resource-rational analysis: understanding human cognition as
  the optimal use of limited computational resources\BBCQ\
\newblock {\Bem Behavioral and Brain Sciences}, {\Bem 43}.

\bibitem[\protect\BCAY{Loreggia, Mattei, Rossi,\ \BBA\ Venable}{Loreggia
  et~al.}{2018a}]{loreggia2018distance}
Loreggia, A., Mattei, N., Rossi, F., \BBA\ Venable, K.~B. \BBOP2018a\BBCP.
\newblock \BBOQ On the distance between {CP}-nets\BBCQ\
\newblock In {\Bem Proc. of the~17th~ AAMAS}, \BPGS\ 955--963. International
  Foundation for Autonomous Agents and Multiagent Systems.

\bibitem[\protect\BCAY{Loreggia, Mattei, Rossi,\ \BBA\ Venable}{Loreggia
  et~al.}{2018b}]{LoMaRoVe18}
Loreggia, A., Mattei, N., Rossi, F., \BBA\ Venable, K.~B. \BBOP2018b\BBCP.
\newblock \BBOQ Preferences and ethical principles in decision making\BBCQ\
\newblock In {\Bem Proc. 1st AIES}.

\bibitem[\protect\BCAY{Loreggia, Mattei, Rossi,\ \BBA\ Venable}{Loreggia
  et~al.}{2018c}]{LoMaRoVe18a}
Loreggia, A., Mattei, N., Rossi, F., \BBA\ Venable, K.~B. \BBOP2018c\BBCP.
\newblock \BBOQ Value alignment via tractable preference distance\BBCQ\
\newblock In Yampolskiy, R.~V.\BED, {\Bem Artificial Intelligence Safety and
  Security}, \BCH~16. CRC Press.

\bibitem[\protect\BCAY{Loreggia, Mattei, Rossi,\ \BBA\ Venable}{Loreggia
  et~al.}{2020a}]{loreggia2020cpmetric}
Loreggia, A., Mattei, N., Rossi, F., \BBA\ Venable, K.~B. \BBOP2020a\BBCP.
\newblock \BBOQ {CPMetric}: Deep siamese networks for metric learning on
  structured preferences\BBCQ\
\newblock In El~Fallah~Seghrouchni, A.\BBACOMMA\  \BBA\ Sarne, D.\BEDS, {\Bem
  Artificial Intelligence. IJCAI 2019 International Workshops}, \BPGS\
  217--234, Cham. Springer International Publishing.

\bibitem[\protect\BCAY{Loreggia, Mattei, Rossi,\ \BBA\ Venable}{Loreggia
  et~al.}{2020b}]{loreggia2020modeling}
Loreggia, A., Mattei, N., Rossi, F., \BBA\ Venable, K.~B. \BBOP2020b\BBCP.
\newblock \BBOQ Modeling and reasoning with preferences and ethical priorities
  in {AI} systems\BBCQ\
\newblock {\Bem Ethics of Artificial Intelligence}, 127.

\bibitem[\protect\BCAY{Mann, Whitney, et~al.}{Mann et~al.}{1947}]{mann1947test}
Mann, H., Whitney, D., et~al. \BBOP1947\BBCP.
\newblock \BBOQ On a test of whether one of two random variables is
  stochastically larger than the other\BBCQ\
\newblock {\Bem Annals of Mathematical Statistics}, {\Bem 18\/}(1), 50--60.

\bibitem[\protect\BCAY{Mikhail}{Mikhail}{2011}]{mikhail2011elements}
Mikhail, J. \BBOP2011\BBCP.
\newblock {\Bem Elements of moral cognition: Rawls' linguistic analogy and the
  cognitive science of moral and legal judgment}.
\newblock Cambridge University Press.

\bibitem[\protect\BCAY{Nichols\ \BBA\ Mallon}{Nichols\ \BBA\
  Mallon}{2006}]{nichols2006moral}
Nichols, S.\BBACOMMA\  \BBA\ Mallon, R. \BBOP2006\BBCP.
\newblock \BBOQ Moral dilemmas and moral rules\BBCQ\
\newblock {\Bem Cognition}, {\Bem 100\/}(3), 530--542.

\bibitem[\protect\BCAY{Noothigattu, Bouneffouf, Mattei, Chandra, Madan,
  Varshney, Campbell, Singh,\ \BBA\ Rossi}{Noothigattu et~al.}{2019}]{NBMC19+}
Noothigattu, R., Bouneffouf, D., Mattei, N., Chandra, R., Madan, P., Varshney,
  K., Campbell, M., Singh, M., \BBA\ Rossi, F. \BBOP2019\BBCP.
\newblock \BBOQ Teaching {AI} agents ethical values using reinforcement
  learning and policy orchestration\BBCQ\
\newblock In {\Bem Proc. of the~28th~IJCAI}.

\bibitem[\protect\BCAY{Parfit}{Parfit}{2011}]{parfit2011matters}
Parfit, D. \BBOP2011\BBCP.
\newblock {\Bem On what matters: volume one}, \lowercase{\BVOL}~1.
\newblock Oxford University Press.

\bibitem[\protect\BCAY{Pu, Faltings, Chen, Zhang,\ \BBA\ Viappiani}{Pu
  et~al.}{2011}]{pu2011usability}
Pu, P., Faltings, B., Chen, L., Zhang, J., \BBA\ Viappiani, P. \BBOP2011\BBCP.
\newblock \BBOQ Usability guidelines for product recommenders based on example
  critiquing research\BBCQ\
\newblock In Ricci, F., Rokach, L., Shapira, B., \BBA\ Kantor, P.~B.\BEDS,
  {\Bem Recommender Systems Handbook}, \BPGS\ 511--545. Springer.

\bibitem[\protect\BCAY{Rawls}{Rawls}{1971}]{rawls1971theory}
Rawls, J. \BBOP1971\BBCP.
\newblock {\Bem A theory of justice}.
\newblock Harvard university press.

\bibitem[\protect\BCAY{Rossi, Venable,\ \BBA\ Walsh}{Rossi
  et~al.}{2011}]{RVW11a}
Rossi, F., Venable, K., \BBA\ Walsh, T. \BBOP2011\BBCP.
\newblock {\Bem A Short Introduction to Preferences: Between Artificial
  Intelligence and Social Choice}.
\newblock Morgan and Claypool.

\bibitem[\protect\BCAY{Rossi\ \BBA\ Loreggia}{Rossi\ \BBA\
  Loreggia}{2019}]{rossi2019preferences}
Rossi, F.\BBACOMMA\  \BBA\ Loreggia, A. \BBOP2019\BBCP.
\newblock \BBOQ Preferences and ethical priorities: {T}hinking fast and slow in
  {AI}\BBCQ\
\newblock In {\Bem Proc. of the~18th~ AAMAS}, \BPGS\ 3--4.

\bibitem[\protect\BCAY{Rossi\ \BBA\ Mattei}{Rossi\ \BBA\ Mattei}{2019}]{RoMa19}
Rossi, F.\BBACOMMA\  \BBA\ Mattei, N. \BBOP2019\BBCP.
\newblock \BBOQ Building ethically bounded {AI}\BBCQ\
\newblock In {\Bem Proc. of the~33rd~AAAI(Blue Sky Track)}.

\bibitem[\protect\BCAY{Russell}{Russell}{2019}]{russell2019human}
Russell, S. \BBOP2019\BBCP.
\newblock {\Bem Human compatible: Artificial intelligence and the problem of
  control}.
\newblock Penguin.

\bibitem[\protect\BCAY{Russell, Dewey,\ \BBA\ Tegmark}{Russell
  et~al.}{2015}]{russell2015research}
Russell, S., Dewey, D., \BBA\ Tegmark, M. \BBOP2015\BBCP.
\newblock \BBOQ Research priorities for robust and beneficial artificial
  intelligence\BBCQ\
\newblock {\Bem AI Magazine}, {\Bem 36\/}(4), 105--114.

\bibitem[\protect\BCAY{Scanlon et~al.}{Scanlon et~al.}{1998}]{scanlon1998we}
Scanlon, T.\BBACOMMA\  et~al. \BBOP1998\BBCP.
\newblock {\Bem What we owe to each other}.
\newblock Harvard University Press.

\bibitem[\protect\BCAY{Sen}{Sen}{1974}]{Sen}
Sen, A. \BBOP1974\BBCP.
\newblock \BBOQ Choice, ordering and morality\BBCQ\
\newblock In K{\"o}rner, S.\BED, {\Bem Practical Reason}. Blackwell, Oxford.

\bibitem[\protect\BCAY{Simon}{Simon}{1955}]{simon1955behavioral}
Simon, H.~A. \BBOP1955\BBCP.
\newblock \BBOQ A behavioral model of rational choice\BBCQ\
\newblock {\Bem The quarterly journal of economics}, {\Bem 69\/}(1), 99--118.

\bibitem[\protect\BCAY{Simon}{Simon}{1956}]{simon1956rational}
Simon, H.~A. \BBOP1956\BBCP.
\newblock \BBOQ Rational choice and the structure of the environment.\BBCQ\
\newblock {\Bem Psychological review}, {\Bem 63\/}(2), 129.

\bibitem[\protect\BCAY{Svegliato, Nashed,\ \BBA\ Zilberstein}{Svegliato
  et~al.}{2021}]{svegliato2021ethically}
Svegliato, J., Nashed, S.~B., \BBA\ Zilberstein, S. \BBOP2021\BBCP.
\newblock \BBOQ Ethically compliant sequential decision making\BBCQ\
\newblock In {\Bem Proceedings of the 35th AAAI International Conference on
  Artificial Intelligence (AAAI)}.

\bibitem[\protect\BCAY{van Baar, Chang,\ \BBA\ Sanfey}{van Baar
  et~al.}{2019}]{van2019computational}
van Baar, J.~M., Chang, L.~J., \BBA\ Sanfey, A.~G. \BBOP2019\BBCP.
\newblock \BBOQ The computational and neural substrates of moral strategies in
  social decision-making\BBCQ\
\newblock {\Bem Nature communications}, {\Bem 10\/}(1), 1--14.

\bibitem[\protect\BCAY{Wallach\ \BBA\ Allen}{Wallach\ \BBA\
  Allen}{2008}]{wallach2008moral}
Wallach, W.\BBACOMMA\  \BBA\ Allen, C. \BBOP2008\BBCP.
\newblock {\Bem Moral machines: Teaching robots right from wrong}.
\newblock Oxford University Press.

\bibitem[\protect\BCAY{Wang, Shao, Zhou, Wan,\ \BBA\ Bouguettaya}{Wang
  et~al.}{2009}]{wang2009web}
Wang, H., Shao, S., Zhou, X., Wan, C., \BBA\ Bouguettaya, A. \BBOP2009\BBCP.
\newblock \BBOQ Web service selection with incomplete or inconsistent user
  preferences\BBCQ\
\newblock In {\Bem Proc. 7th International Conference on Service-Oriented
  Computing}, \BPGS\ 83--98. Springer.

\bibitem[\protect\BCAY{Yu, Shen, Miao, Leung, Lesser,\ \BBA\ Yang}{Yu
  et~al.}{2018}]{yu2018building}
Yu, H., Shen, Z., Miao, C., Leung, C., Lesser, V.~R., \BBA\ Yang, Q.
  \BBOP2018\BBCP.
\newblock \BBOQ Building ethics into artificial intelligence.\BBCQ\
\newblock In {\Bem Proc. 27th IJCAI}, \BPGS\ 5527--5533.

\end{thebibliography}
